\documentclass[11pt,a4paper]{article}
\usepackage[hyperref]{emnlp2020}
\usepackage{times}
\usepackage{latexsym}

\usepackage{microtype}

\aclfinalcopy 
\def\aclpaperid{668} 

    \usepackage{url}
    \usepackage{hyperref}
    \usepackage{longtable}
    \usepackage{makecell}
    \usepackage{fontawesome}
    \usepackage{setspace}
    \usepackage{graphicx}
    \usepackage{booktabs}
    \usepackage{amsmath}
    \usepackage{multirow}
    \usepackage{multicol}
    \usepackage{tabularx}
    \usepackage{enumitem}
    \usepackage{collcell}
    \usepackage{color, colortbl}
    \usepackage{cleveref}
    \usepackage{arydshln}
    \usepackage{amssymb}
    \usepackage{array}
    \newcolumntype{H}{>{\setbox0=\hbox\bgroup}c<{\egroup}@{}}
    
    \crefformat{section}{\S#2#1#3}
    \crefformat{subsection}{\S#2#1#3}
    \crefformat{subsubsection}{\S#2#1#3}
    \crefrangeformat{section}{\S\S#3#1#4 to~#5#2#6}
    \crefmultiformat{section}{\S\S#2#1#3}{ and~#2#1#3}{, #2#1#3}{ and~#2#1#3}
    \newcommand*{\mymathrm}[1]{$\mathrm{#1}$}
    \usepackage{tcolorbox}
    \definecolor{tagblue}{HTML}{6374d6}
    \definecolor{tagyellow}{HTML}{ffea9e}
    
    \makeatletter
    \newtcbox{\bluetag}{valign=center, colframe=tagblue,colback=tagblue!50, boxrule=0pt,arc=2pt, boxsep=0pt,left=4pt,right=4pt,top=0pt,bottom=0pt, nobeforeafter, height=10pt, enlarge bottom by=-3pt, fontupper=\ttfamily}
    
    \newtcbox{\yellowtag}{valign=center, colframe=tagyellow,colback=tagyellow!40, boxrule=0pt,arc=2pt, boxsep=0pt,left=4pt,right=4pt,top=0pt,bottom=0pt, nobeforeafter, height=10pt, enlarge bottom by=-3pt, fontupper=\ttfamily}

    \title{Multi-view Story Characterization from 
    Movie Plot Synopses and Reviews}
    
    \newtoggle{human_eval}
    \toggletrue{human_eval}
    
    \author{Sudipta Kar$^\clubsuit$, Gustavo Aguilar$^\clubsuit$, Mirella Lapata$^\spadesuit$, Thamar Solorio$^\clubsuit$\\
        $^\clubsuit$ University of Houston\\
      $^\spadesuit$ ILCC, University of Edinburgh\\
      \texttt{\{skar3, gaguilaralas\}@uh.edu, mlap@inf.ed.ac.uk, tsolorio@uh.edu}}
    
    \date{}
    
\begin{document}
    \maketitle

    \begin{abstract} 
    This paper considers the problem of characterizing stories by inferring properties such as theme and style using written synopses and reviews of movies.
We experiment with a multi-label dataset of movie synopses and a tagset representing various attributes of stories (e.g., genre, type of events).
Our proposed multi-view model encodes the synopses and reviews using hierarchical attention and shows improvement over methods that only use synopses.
Finally, we demonstrate how can we take advantage of such a model to extract a complementary set of story-attributes from reviews without direct supervision.
We have made our dataset and source code publicly available at \url{https://ritual.uh.edu/multiview-tag-2020}.

    \end{abstract}
    
    \section{Introduction}
    A high-level description of stories represented by a tagset can assist consumers of story-based media (e.g., movies, books) during the selection process.
Although collecting tags from users is time-consuming and often suffers from coverage issues \cite{katakis-incompleteness}, NLP techniques like those in \newcite{folksonomication2018} and \newcite{gorinski_naacl_N18-1160} can be employed to generate tags automatically from written narratives such as synopses.
However, existing supervised approaches suffer from two significant weaknesses. Firstly, the accuracy of the extracted tags is subject to the quality of the synopses. Secondly, the tagset is predefined by what was present in the training and development sets and thus is brittle; story attributes are unbounded in principle and grow with the underlying vocabulary.

\begin{figure}[t]
\centering
\resizebox{\columnwidth}{!}{%
\begin{tabular}{c}
\textbf{\fontfamily{ppl} \small{Plot Synopsis}}\\
\multicolumn{1}{p{\columnwidth}}{\fontfamily{cmss} \scriptsize
 \cellcolor{tagblue!20}In late summer 1945, guests are gathered for the wedding reception of Don Vito Corleone's daughter Connie (Talia Shire) and Carlo Rizzi (Gianni Russo). ... ... The film ends with Clemenza and new caporegimes Rocco Lampone and Al Neri arriving and paying their respects to Michael. Clemenza kisses Michael's hand and greets him as "Don Corleone." As Kay watches, the office door is closed."}\\
\textbf{\fontfamily{ppl} \small{Review}}\\
\multicolumn{1}{p{\columnwidth}}{\fontfamily{cmss} \scriptsize
\cellcolor{tagyellow!25} Even if the viewer does not like mafia type of movies, he or she will watch the entire film ... ... Its about family, loyalty, greed, relationships, and real life. This is a great mix, and the artistic style make the film memorable.}\\\hline



\bluetag{\small violence} \bluetag{\small action} \bluetag{\small murder}  \bluetag{\small atmospheric}\\  \bluetag{\small revenge} 
\yellowtag{\small mafia} \yellowtag{\small family} \yellowtag{\small loyalty}\\  \yellowtag{\small greed} \yellowtag{\small relationship} \yellowtag{\small artistic} 
\\

\end{tabular}%
}
\caption{\small Example snippets from plot synopsis and review of {\fontfamily{ppl} \textbf{The Godfather}} and tags that can be generated from these.}
\label{tab:front_page_example_table}
\vspace{-1em}
\end{figure}

To address the weaknesses presented above, we propose to exploit user reviews. We have found that movie reviews often discuss many aspects of the story.
For example, in Figure \ref{tab:front_page_example_table}, a reviewer writes that {\fontfamily{ppl} \selectfont{\textit{The Godfather}}} is about \textit{family, relations, loyalty, greed}, and \textit{mafia}, whereas the gold standard tags from the plot are \textit{violence, murder, atmospheric, action}, and \textit{revenge}. 
In this paper, we show that such information in reviews can significantly strengthen a supervised \textit{synopses to tag} prediction system, hence alleviating the first limitation.
To address the second limitation, we propose to also rely on the content provided by the reviews.
We extract new tags from reviews and thus complement the predefined tagset.

A potential criticism of an approach that relies on user reviews is that it is not practical to wait for user reviews to accumulate. The speed at which movies get reviews fluctuates a lot.
Therefore, we propose a system that learns to predict tags by jointly modeling the movie synopsis and its reviews, when the reviews are available. But if a movie has not accumulated reviews yet, our system can still predict tags from a predefined tagset using only the synopsis without any configuration change.

Tag extraction from reviews can be modeled as a supervised aspect extraction problem \cite{liu2012sentiment} that requires a considerably large amount of annotated tags in the reviews. 
To get rid of this annotation burden, we formulate the problem from the perspective of Multiple Instance Learning (MIL; \citeauthor{mil-keeler1992}, \citeyear{mil-keeler1992}).
As a result, our model learns to spot story attributes in reviews in a weakly supervised fashion and does not expect direct tag level supervision. 
Note that these complementary open-vocabulary tags extracted from the reviews are separate from the predefined tagset, and we can only generate this set when a movie has reviews.
As we show in Section \ref{sec:human_eval_results}, tags generated by our system can quickly describe a story helping users decide whether to watch a movie or not.

Our contributions in this work can be summarized as follows:
\begin{itemize}[]
    
    \item We collect $\approx$1.9M  user reviews to enrich an existing dataset of movie plot synopses and tags.
    
    \item We propose a multi-view multi-label tag prediction system that learns to predict relevant tags from a predefined tagset by exploiting the synopsis and the reviews of a movie when available. We show that utilizing reviews can provide $\approx$4\% increase in F1 over a system using only synopses to predict tags.
    
    \item We demonstrate a technique to extract open-vocabulary descriptive tags (i.e., not part of the predefined tagset) from reviews using our trained model.
    While review-mining is typically approached as supervised learning, we push this task to an unsupervised direction to avoid the annotation burden.

\end{itemize}
 We verify our proposed method against multiple competitive baselines and conduct a human evaluation to confirm our tags' effectiveness for a set of movies.
    
    \section{Background}
    Prior art related to this paper's work includes story analysis of movies and mining opinions from movie reviews. In this section, we briefly discuss these lines of work.

\paragraph{Story Analysis of Movies}
Over the years, high-level story characterization approaches evolved around the problem of identifying genres \cite{Biber_Douglas1992, genre_E97-1005, cross_lingual_genre_E12-3002, genre_identification_C18-1167}. 
Genre information is helpful but not very expressive most of the time as it is a broad way to categorize items. 
Recent work \cite{gorinski_naacl_N18-1160, folksonomication2018} retrieves other attributes of movie storylines like \textit{mood}, \textit{plot type}, and \textit{possible feeling of consumers} in a supervised fashion where the number of predictable categories is more extensive and more comprehensive compared to genre classification.
Even though these systems can retrieve comparatively larger sets of story attributes, the predictable attributes are limited in a closed group of tags. In contrast, in real life, these attributes can be unlimited.

\paragraph{Movie Review Mining}
There is a subtle distinction between the reviews of typical material products (e.g. \textit{phone, TV, furniture)} and story-based items (e.g. \textit{literature, film, blog}).
In contrast to the usual aspect based opinions (e.g. \textit{battery, resolution, color}),  reviews of story-based items often contain \textit{end users' feelings, important events of stories}, or \textit{genre related information}, which are abstract in nature (e.g. \textit{heart-warming, slasher, melodramatic}) and do not have a very specific target aspect.
Extraction of such opinions about stories has been approached by previous work using reviews of movies \cite{ movie_review_mining_Zhuang, structure_review_mining_C10-1074} and books \cite{book_review_mining}.
Such attempts are broadly divided into two categories.
The first category deals with spotting words or phrases (\textit{excellent, fantastic, boring}) used by people to express how they feel about the story. And the second category focuses on extracting important opinionated sentences from reviews and generating a summary.
In our work, while the primary task is to retrieve relevant tags from a pre-defined tagset by supervised learning, our model provides the ability to mine story aspects from reviews without any direct supervision.
    
    \section{Dataset}\label{sec:dataset}
    \begin{table}[t]
\centering
\small
\setlength\belowcaptionskip{-18pt}
\begin{tabular}{
				l 
				>{\collectcell\mymathrm}c<{\endcollectcell} >{\collectcell\mymathrm}c<{\endcollectcell}
				>{\collectcell\mymathrm}c<{\endcollectcell}}
\hline
& \mathbf{Train} & \mathbf{Val} & \mathbf{Test}\\ \hline
Instances & 9,746 & 2,437 & 3,046\\
Tags per instance & 3 & 3 & 3 \\
Reviews per movie & 72 & 74 & 72\\

$^S$ Sentence per document   & 50 & 53 & 51 \\
$^S$ Words per sentence & 21 & 21 & 21 \\

$^R$ Sentence per document  & 117 & 116 & 116 \\
$^R$ Words per sentence & 27 & 27 & 27\\

\hline
\end{tabular}
\caption{\small Statistics of the dataset. $^S$ denotes synopses and $^R$ denotes review summaries.}
\label{tab:data_stat}
\end{table}

Our starting data set is the MPST corpus \cite{mpst2018} which contains approximately 15K movies and a set of tags assigned by  IMDB\footnote{\url{http://imdb.com}} and MovieLens\footnote{\url{http://movielens.org}} users.
The tagset contains 71 labels that are representative of story related attributes (e.g., \textit{thought-provoking, inspiring, violence}) and the corpus is free from any metadata (e.g., \textit{cast, release year}).

We extended the dataset by collecting up to 100 most helpful reviews per movie from IMDB.
Out of the 15K movies in MPST, we did not find any reviews for 285 films.
The collected reviews often narrate the plot summary and describe opinions about movies.
We noticed that reviews can be very long and sometimes contain repetitive plot summaries and opinions.
Some reviews can be even uninformative about the story type.
Moreover, the number of reviews for movies varies greatly, creating a challenge for modeling them computationally.
So we summarize all reviews for a movie into a single document using TextRank\footnote{We used the implementation from the Gensim library and converted the reviews into Unicode before summarizing.} \cite{textrank2004}.
We observed that summarized reviews are usually free of repetitive information and aggregate the salient fragments from the reviews that are heavy with user opinions.
All plot synopses, reviews, and tags are in English, and Table \ref{tab:data_stat} presents some statistics of the dataset.\footnote{Some example plot synopses and reviews are presented in Appendix \ref{app:example_plot_review}.}

    \section{Modeling}\label{sec:modeling}
\begin{figure*}
    \centering
    \includegraphics[width=\textwidth]{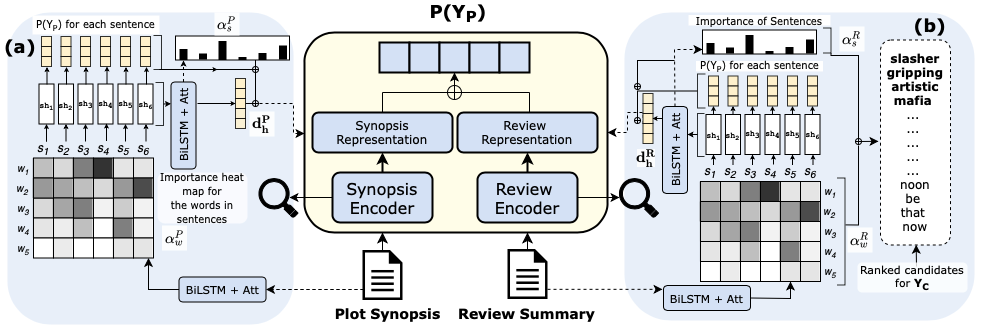}
    \caption{\small In the center, we show an overview of the model that takes a plot synopsis and a review summary as input, uses two separate encoders to construct the high-level representations and uses them to compute $P(Y_P)$.
    \textbf{(a)} illustrates an enhanced view of the synopsis encoder. It uses a BiLSTM with attention to compute a representation $\mathbf{sh_i^P}$ for the $i_{th}$ sentence in the synopsis. 
    Additionally, a matrix of word-level attention weights $\alpha_w^P$ is generated that indicates the importance of each word in each sentence for correctly predicting $P(Y_P)$.
    Another attention based BiLSTM is used to create a synopsis representation $d_h^P$ from the encoded sentences $\mathbf{sh^P}$.
    Additionally, for each $\mathbf{sh_i^P}$, sentence-level prediction $P(Y_P)_i$ is computed which is aggregated with $\mathbf{d_h^P}$ to create the final synopsis representation.
    \textbf{(b)} illustrates a similar encoder for reviews.
    To create a complementary tagset $Y_C$ by mining the reviews, word-level importance scores $\alpha_w^R$ and sentence-level importance scores $\alpha_s^R$ are used (Equation \ref{eq:review_tag}). 
    Apart from that, review representation $\mathbf{d_h^R}$ is computed in a similar way as in \textbf{(a)} which is used together with $\mathbf{d_h^P}$ to compute $P(Y_P)$.
    }
    \label{fig:model}
\end{figure*}

    Consider  input $X=\{X_{PS}, X_R\}$, where $X_{PS}$ is a plot synopsis and $X_{R}$ is a review summary. For a predefined tagset $Y_P = [y_1, y_2, ..., y_{|Y_P|}]$, we want to model $P(Y_P|X)$.
We also aim at extracting a complementary tagset $Y_C$ from $X_R$ that is not part of the original $Y_P$ set and is not labeled in the dataset.
However, we expect a latent correlation between $Y_P$ and $Y_C$ that can be jointly modeled while modeling $P(Y_P|X)$, hence helping the extraction of $Y_C$ without any direct supervision.
Therefore, we first supervise a model containing a synopsis encoder and a review encoder to learn $P(Y_P|X)$ (Section \ref{sec:model_encoding}), and later we use the trained review encoder to generate complementary tagset $Y_C$  (Section \ref{sec:review_tags}).
An overview of our model is shown in Figure \ref{fig:model}.

\subsection{Learning the Predefined Tagset}\label{sec:model_encoding}
Different words and sentences in a synopsis have different roles in the overall story. For example, some sentences narrate the setting or background of a story, whereas other sentences may describe different events and actions. Additionally, some sentences and words are more helpful for identifying relevant tags from the synopsis.
With this in mind, we adapt the hierarchical encoding technique of \newcite{han_yang-EtAl:2016:N16-13} that learns to weight important words and sentences and use this information to create a high-level document representation.
Additionally, to efficiently capture various important story aspects in long synopses and reviews, we model our task from the perspective of Multiple Instance Learning (MIL).

We assume that each synopsis and review is a bag of instances (i.e., sentences in our task), where labels are assigned at the bag level. 
In such cases, a prediction is made for the bag by either learning to aggregate the instance level predictions \cite{mil-keeler1992, milDietterich, mil-Maron98multiple-instancelearning} or jointly learning the labels for instances and the bag \cite{mil-Zhou09multi-instancelearning,mil-wei,mil-Kotzias2015FromGT,mil-angelidis-lapata-2018-multiple,mil-yumo}.
In our setting, we choose the latter; i.e., we aggregate $P(Y_P)$ for each sentence with the combined representation of $X_{PS}$ and $X_R$ to compute $P(Y_P|X)$.
As we will show later, MIL improves prediction performance and promotes interpretability.

We represent a synopsis $X_{PS}$ consisting of $L$ sentences  $(s_{1}, ..., s_{L})$ in a hierarchical manner instead of a long sequence of words.
At first, for a sentence $s_i = (w_{1}, ..., w_{T})$ having $T$ words, we create a matrix $E_i$ where $E_{it}$ is the vector representation for word $w_{t}$ in $s_i$.
We use pre-trained Glove embeddings \cite{pennington2014glove} to initialize $E$.
Then, 
we encode the sentences using a bidirectional LSTM  \cite{lstm:hochreiter:1997}  with attention \cite{attentionBahdanauCB14}.
It helps the model to create a sentence representation $\mathbf{sh}_i $ for the $i_{th}$ sentence in $X_{PS}$ by learning to put a higher weight on the words that correlate more with the target tags.
The transformation is as follows:
\begin{flalign*}
    \quad &\overrightarrow{h}_{w_{it}} = \overrightarrow{LSTM}(\mathbf{E}_{it}), t \in [1, T] \\ 
    \quad &\overleftarrow{h}_{w_{it}} = \overleftarrow{LSTM}(\mathbf{E}_{it}), t \in [T, 1] \\
    \quad &\mathbf{u}_{it} = \tanh(\mathbf{W}_{wt} . [\overrightarrow{h}_{w_{it}},  \overleftarrow{h}_{w_{it}}] + \mathbf{b}_w) &\\
    \quad &\mathbf{r}_{it} = \mathbf{u}_{it}^\top \mathbf{v}_t; \quad
    \mathbf{\alpha}_{it} = \frac{\exp(\mathbf{r}_t)}{\sum_{t}\exp(\mathbf{r}_t)}\\
    \quad &\mathbf{sh}_{i} = \sum_{t=1}^{T} \mathbf{\alpha}_{it}  \mathbf{h}_{it}
\end{flalign*}

Here, $\mathbf{W}_{wt}, \mathbf{b}_w, \mathbf{v}_t$ are learned during training.
In the second step, we pass the encoded sentences $\mathbf{sh}$ through another BiLSTM layer with attention.
By taking the weighted sum of the hidden states and attention scores $\alpha_s^{PS}$ for the sentences, we generate an intermediate document representation $\mathbf{d_h^{PS'}}$.
Simultaneously, for each high level sentence representation $\mathbf{sh}_i$, we predict $P(Y_P)_{s_i}^{PS}$.
We then weight $P(Y_P)_{s_i}^{PS}$ by $\alpha_s^{PS}$ and compute a weighted sum to prioritize the predictions made from comparatively important sentences.
This sum is aggregated with $\mathbf{d_h^{PS}}$ to generate the final document representation.
\paragraph{Aggregating Synopses and Reviews}
After generating the high level representation of the synopses ($\mathbf{d_h^{PS}}$) and reviews ($\mathbf{d_h^R}$), we merge them to predict
\begin{align*}
    P(Y_P) = \textit{Softmax}(\mathbf{W}_o \cdot [\mathbf{d_h^{PS}, d_h^R}] + b_o)
\end{align*}
Here, $\mathbf{W}_o$ and $b_o$ are learnable weight and bias of the output layer (dimension=$|Y_P|$, i.e.,71), respectively.
We experiment with two types of aggregation techniques: a) simple concatenation, and b) gated fusion.
In the first approach, we concatenate these two representations, whereas in the second approach, we control the information flow from the synopses and reviews.
While important story events and settings found in synopses can correlate with some tags, viewers' reactions can also correlate with complementary tags.
We believe that learning to control the contribution of information encoded from synopses and reviews can improve overall model performance.
For instance, if the synopsis is not descriptive enough to retrieve relevant tags, but the reviews have adequate information, we want the model to use more information from the reviews.
Hence, we use a gated fusion mechanism \cite{arevalo2017gated} on top of the encoded synopsis and review representations. 
For the encoded synopsis $\mathbf{d_h^{PS}}$ and review representation $\mathbf{d_h^R}$, the mechanism works as follows:
\begin{flalign*}
\quad  h_{ps} &= \tanh(\mathbf{W}_{ps} \cdot \mathbf{d_h^{ps}})\\
h_r &= \tanh(\mathbf{W}_r \cdot \mathbf{d_h^r})\\
\quad  z &= \sigma(\mathbf{W}_z \cdot [\mathbf{d_h^{ps}}, \mathbf{d_h^r}])\\
\quad  h &= z * h_{ps} + (1 - z) * h_r
\end{flalign*}

\subsection{Tag Generation from Reviews}\label{sec:review_tags}
Extracting tags from reviews can be seen from the perspective of MIL, where instance (i.e., word) level annotations are not present, but each movie is labeled with some tags from the predefined tagset $Y_P$.
When we train the model (Section \ref{sec:model_encoding}), these bag-level labels seem to act as weak supervision for the model to learn to isolate instances --- i.e., tags present in the reviews.
For example, we observe that the model usually puts higher attention weights on opinion--heavy words in the reviews.
Therefore, we use the attention weights on words and sentences in reviews to extract an additional open-vocabulary tagset $Y_C$.

\begin{figure}
    \centering
    \includegraphics[width=\columnwidth]{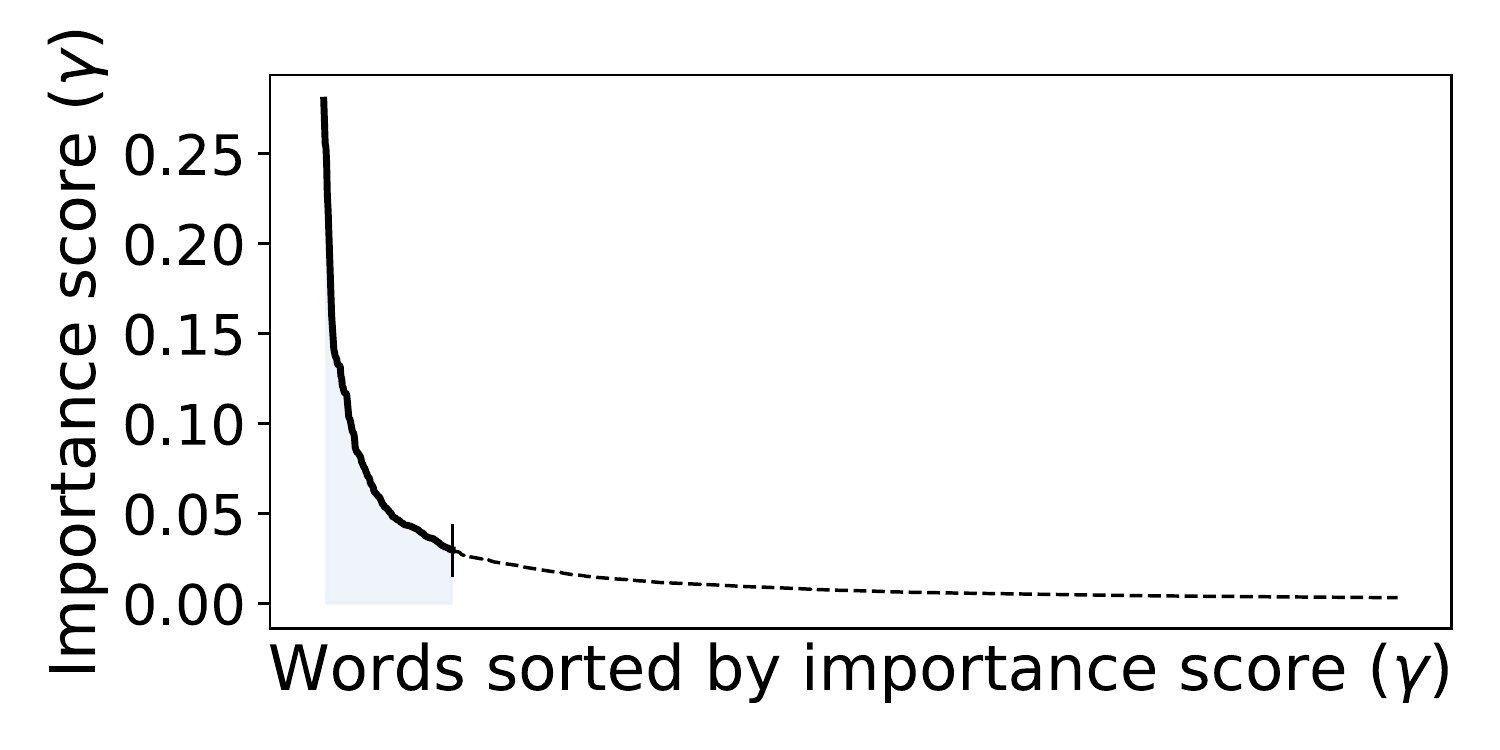}
    \caption{\small All words in the shaded area under the solid black curve are selected as candidate tags for $Y_C$.}
    \label{fig:slope}
\end{figure}

Predicting $P(Y_P)$ from $\{X_{PS}, X_R\}$ produces attention weight vectors $\alpha_{W}^R$ and $\alpha_{s}^R$ for $X_{R}$ (as in Section \ref{sec:model_encoding}).
For each word $w_{ij}$ in sentence $s_i$ in $X_{R}$, we compute an importance score $\gamma_{ij}$ as:
\begin{align}
    \gamma_{ij} = \alpha_{W_{ij}} \times \alpha_{s_i} \times  |s_i|
    \label{eq:review_tag}
\end{align}
Here, $\alpha_{W_{ij}}$ is the attention weight of word $w_{ij}$ and $\alpha_{s_i}$ is the attention weight of the $i^{th}$ sentence. 
$|s_i|$ indicates the number of words in the sentence, and helps overcome the fact that word-level attention scores are higher in shorter sentences.
We rank the words in the reviews based on their importance scores and choose the first few words as the primary candidates for $Y_C$ as shown in Figure-\ref{fig:slope}.
The idea is that the sorted scores create a downward slope, which allows us to stop at the point where the slope starts getting flat.
We detect this by computing the derivative at this point based on its neighboring four points and set a threshold of $5e^{-3}$ based on our observations on the validation set.
After selecting the candidates, we remove duplicates and tags that are already in the predefined tagset $Y_P$ to avoid redundancy.
This method gives us a new open-vocabulary tagset $Y_C$ created from the reviews without any direct supervision.

    \section{Experiments}
    We treat our tag assignment task as a multi-label classification problem.
Based on $P(Y_P | X)$, we sort the predefined tagset $Y_P$ in descending order, so that tags with higher weights are ranked on top.
Then, in different settings we select the \textit{top-k} (\textit{k}=3, 5) tags as the final tags to describe each movie.
We aim to explore three research questions through our experiments: (Q1) for predicting tags from synopses only, can our approach outperform other machine learning models? (Q2) When available, can reviews strengthen the \textit{synopses to tag} prediction model? and (Q3) how relevant are open-vocabulary tags to stories?

For Q1 and Q2, we evaluate systems based on two aspects: a) correctness of \textit{top-k} predictions by micro-F1, and b) diversity in \textit{top-k} tags using Tags Learned (TL; \citeauthor{mpst2018}, \citeyear{mpst2018}). TL is simply the number of unique tags predicted for the entire evaluation set in \textit{top-k} setting; i.e., $|Y^k_{P_{pred}}|$.
We verify Q3 through a human evaluation experiment as we do not have annotations for review tags.

\subsection{Baselines}
We compare our model against the following baselines:
    \paragraph{Most Frequent} Most frequent $k (3, 5)$ tags from the predefined tagset $Y_P$ are assigned to each movie. \vspace{-1em}
    
    \paragraph{Convolutional Neural Network with Emotion Flow (CNN-EF)} We use a Convolutional neural network-based text encoder to extract features from written synopses and Bidirectional LSTMs to model the flow of emotions in the stories \cite{folksonomication2018}. To our knowledge, this method is currently the best-performing system on our task.
    
    \paragraph{Pre-trained language models} Large pre-trained language models (LM) built with Transformers \cite{transformers} have shown impressive performance in a wide range of natural language understanding (NLU) tasks like natural language inference, sentiment analysis, and question-answering in the GLUE benchmark \cite{wang2019glue}.
    However, directly fine-tuning such models for long texts like synopses and reviews is extremely memory expensive.
    Therefore, we employ Sentence-BERT (SBERT; \citeauthor{reimers-gurevych-2019-sentence}, \citeyear{reimers-gurevych-2019-sentence}) in our work, which is a  state-of-the-art universal sentence encoder built with pre-trained BERT \cite{devlin-etal-2019-bert}.
    We use SBERT encoded sentence representations with our proposed model in Section \ref{sec:modeling} instead of training the Bi-LSTM with a word-level attention based sentence encoder.
    Then we use these representations to create a document representation using Bi-LSTM with sentence-level attention, keeping the rest of the model unchanged.

    \section{Results}\label{sec:results}
    \paragraph{Quantitative Results} We report the results of our experiments on the test\footnote{Validation results are provided in Appendix \ref{app:val_results}.} set in Table \ref{tab:test_results}.
We mainly discuss the \textit{top-3} setting, where three tags are assigned to each instance by all systems.

Regarding our first research question, Table \ref{tab:test_results} shows that our proposed hierarchical model with attention HN(A) outperforms all comparison systems (F1=37.90).
This model achieves slightly better F1 than SBERT, which implies that word-level attention must be learned for accurate tag prediction.
Additionally, learning document level attention is also crucial as HN(A) performs better than using maxpool.
Finally, sentence level tag prediction (\textit{HN(A) + MIL row}) is beneficial for both accurate and diverse tagset generation (F1=37.94, TL=51).
\begin{table}[t]
\centering
\resizebox{0.9\columnwidth}{!}{%
\begin{tabular}{>{\collectcell\mymathrm}l<{\endcollectcell}
H
>{\collectcell\mymathrm}c<{\endcollectcell} >{\collectcell\mymathrm}c<{\endcollectcell}
>{\collectcell\mymathrm}c<{\endcollectcell} >{\collectcell\mymathrm}c<{\endcollectcell}}
\hline

& & \multicolumn{2}{c}{{$\mathrm{\mathbf{Top-3}}$}} & \multicolumn{2}{c}{\textbf{$\mathrm{\mathbf{Top-5}}$}}\\
\cmidrule(lr){3-4}
\cmidrule(lr){5-6}

\multicolumn{1}{c}{} & \multicolumn{1}{H}{{MLR}} &\multicolumn{1}{c}{$\mathrm{\mathbf{F1}}$} & \multicolumn{1}{c}{$\mathrm{\mathbf{TL}}$} & \multicolumn{1}{c}{$\mathrm{\mathbf{F1}}$} & \multicolumn{1}{c}{$\mathrm{\mathbf{TL}}$}\\ \hline

\multicolumn{6}{l}{Synopsis to Tags}\\ \hdashline
Most \quad Frequent  & 85.66 & 29.70 & 3 & 28.40 & 5 \\
CNN-EF  & 86.51 & 36.90 & \mathbf{58} & 36.70 & \mathbf{65}\\ 
SBERT & 90.54 & 37.44 & 39 & 37.38 & 46\\

HN (Maxpool) & 89.32 & 36.31 & 17 & 36.01 & 26\\
HN (A) & 90.45 & 37.90 & 37 & 37.67 & 46\\
HN (A) + MIL & \textbf{91.06} & \mathbf{37.94} & {51} & \mathbf{38.25} & {55}\\

\hline 
\multicolumn{6}{l} {Synopsis + Review to Tags}\\ \hdashline
Merge \quad Texts & 92.67 & 41.26 & 51 & 41.11 & 58 \\
Concat \quad Representations & 92.68 & 40.64 & 55 & 40.82 & 62 \\
Gated \quad Fusion^* & \textbf{93.41} & \mathbf{41.84} & \mathbf{64} & \mathbf{41.80} & \mathbf{67} \\

\hdashline 
\multicolumn{6}{l}{Subset: Every movie has at least one review} \\ \hdashline
Synopsis & 91.05 & 38.24 & 50 & 37.99 & 55 \\
Review & 93.51 & 42.00 & 60 & 42.19 & 64\\
Both & \textbf{93.52} & \textbf{42.11} & \textbf{65} & \textbf{42.00} & \textbf{68}\\

\hline
\end{tabular}
}
\caption{\small Results obtained on the test set using different methodologies on the synopses and after adding reviews with the synopses. TL stands for \textit{tags learned}. $^*$: t-test with $p$-value $< 0.01$.}
\label{tab:test_results}
\end{table}

Table \ref{tab:test_results} also shows that reviews combined with synopses can boost tag prediction performance (Q2).
As a simple baseline technique to integrate reviews with synopses, we merge the review texts with synopses to train a single encoder based model.
This technique shows improvements over the model that uses only synopses (F1=41.26).
Using two separate encoders for synopses and reviews, concatenating the generated representations decreases F1 (40.64), but increases TL (55).
Combining these representations by gated fusion achieves the best results so far (F1=41.84, TL=64).
By performing a t-test, we found that gated fusion is significantly better ($p$-value $<0.01$) than merging the texts and simple concatenation of the high-level representations of synopses and reviews.

As we do not have reviews for $\approx$300 movies (Section \ref{sec:dataset}), we further experiment to verify Q2 on a subset of our data, where every movie has at least one review.
As shown in Table \ref{tab:test_results}, reviews act as a stronger data source than synopses for classifying tags.
Combining both does not affect F1 ($\approx$42) much, but TL improves by a considerable margin (60 vs. 65).
We found that combining synopses helps to identify tags like \textit{plot twist, bleak, grindhouse film,} and \textit{allegory}.
It shows that our model is successfully capturing different story attributes from reviews that are possibly difficult to find in synopses.
Again, as reviews are not always available for movies, treating synopses as the primary data source and reviews as complementary information is practical.

\begin{figure}
    \centering
    \includegraphics[width=\columnwidth]{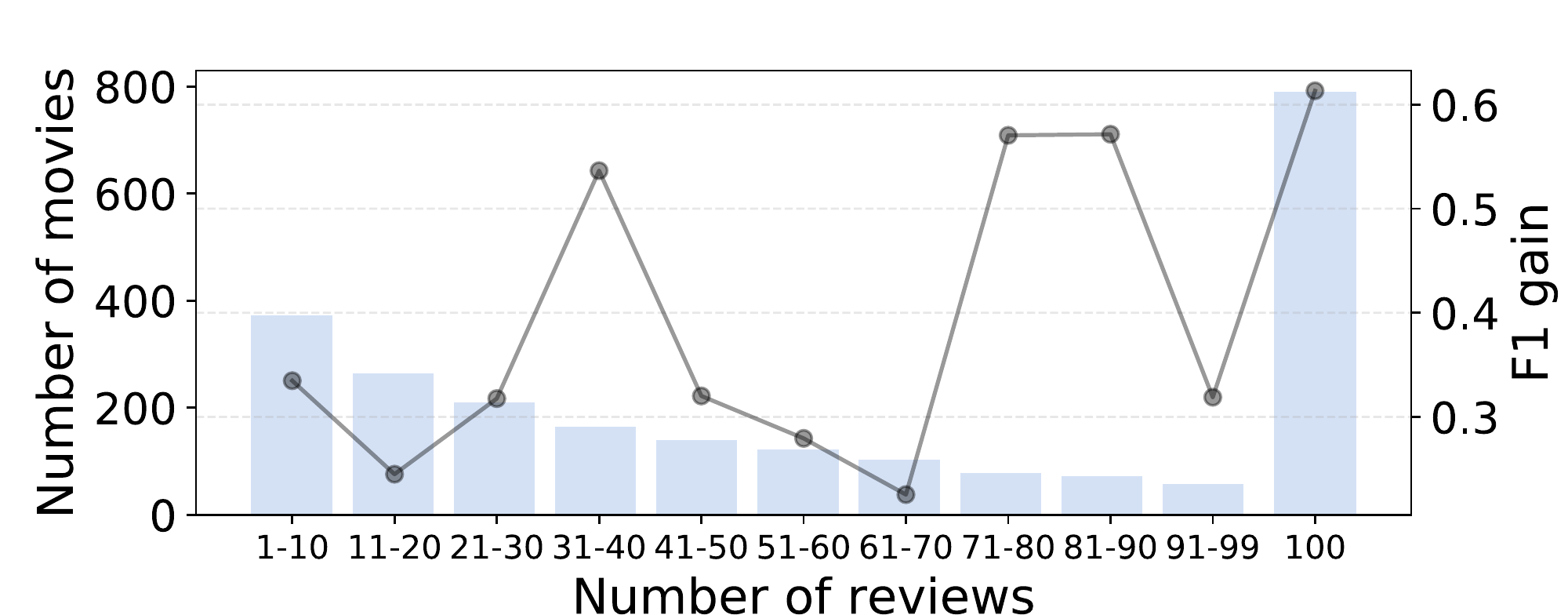}
    \caption{\small Average change in F1 with respect to the number of reviews after combining review summaries with synopses.}
    \label{fig:f1change_nreviews}
\end{figure}

\paragraph{How Many Reviews Do We Need?} 
We investigate the least amount of reviews we require to observe reasonable performance gains.
The curve in Figure \ref{fig:f1change_nreviews} shows that we can expect a noticeable improvement in tag prediction performance if we have at least around 31-40 reviews for a movie.
However, as the plot shows, having less than that can still provide some benefit.
Note that we generate a single summary document from these reviews to feed into the model.
The gain fluctuates for movies having more than 40 reviews and less than 99. This is also the group with the smaller number of movies, so any conclusions for this range should be taken with a grain of salt. However, 790 movies have 100 reviews, and the average gain is slightly better than what we observe with 31-40 reviews.

To better understand the reason behind sudden drops in performance in different bins, we looked at the bins' genre information. 
In IMDb, a movie is generally labeled with multiple genres. 
We observed that movies in bins with higher F1 usually have more gold labels for genres and tags than movies in bins with lower F1. This fact alone, of having more gold tags assigned to the movies, makes it more likely that system prediction tags will match some of them.
And the opposite happens in bins with lower F1.
Additionally, while looking at genres, we found that some less frequent genres like film-noir are comparatively more in bins like 51-60, which can also create a performance gap.

\begin{figure}[t]
    \centering
    \includegraphics[width=\columnwidth]{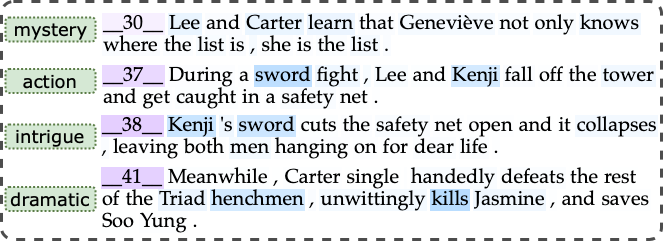}
    \caption{\small Example sentences from the synopsis of the movie {\fontfamily{ppl}{\textbf{Rush Hour 3}}} with one of the most relevant tags from the sentence-level predictions. Importance of particular sentences and words for predicting tags is indicated by the highlight intensity of the sentence ids and words. Ground truth tags are \textit{bleak, violence, comedy, murder}.}
    \label{fig:plot_att}
\end{figure}

\begin{figure}[t]
    \centering
    \includegraphics[width=\columnwidth]{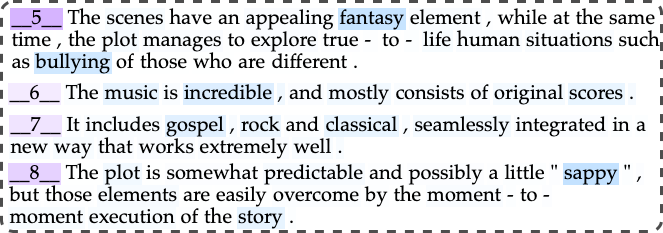}
    \caption{\small Example sentences from the review of the movie {\fontfamily{ppl}{\textbf{August Rush}}} with sentence ids and words highlighted based on their importance in tag prediction. Ground truth tags are \textit{thought-provoking, romantic, inspiring, flashback}.}
    \label{fig:review_att}
\end{figure}

\paragraph{Are These Hierarchical Representation Meaningful? }
We analyze the reason behind the effectiveness of our proposed system by visualizing the attention weights at the sentence and word level for the synopsis of \textit{Rush Hour 3} and the reviews of \textit{August Rush} (see Figure \ref{fig:plot_att} and \ref{fig:review_att}, respectively).
We can see that sentences in the synopsis that describe important story events and sentences in the review that express user opinions about the story receive higher weights.
Similarly, at the sentence level, important events and characters are weighted more by the model, and words in review sentences that convey opinions about the storyline rather than other aspects of the movie (e.g., music) receive more weight by the model.
If we observe the tagsets provided in the caption of Figure \ref{fig:review_att} and the highlighted words and sentences, we can conclude that the model is efficiently modeling the correlations between salient parts of the text and tags.

\subsection{Human Evaluation}\label{sec:human_eval_results}
\begin{figure}[t]
    \centering
    \includegraphics[width=\columnwidth]{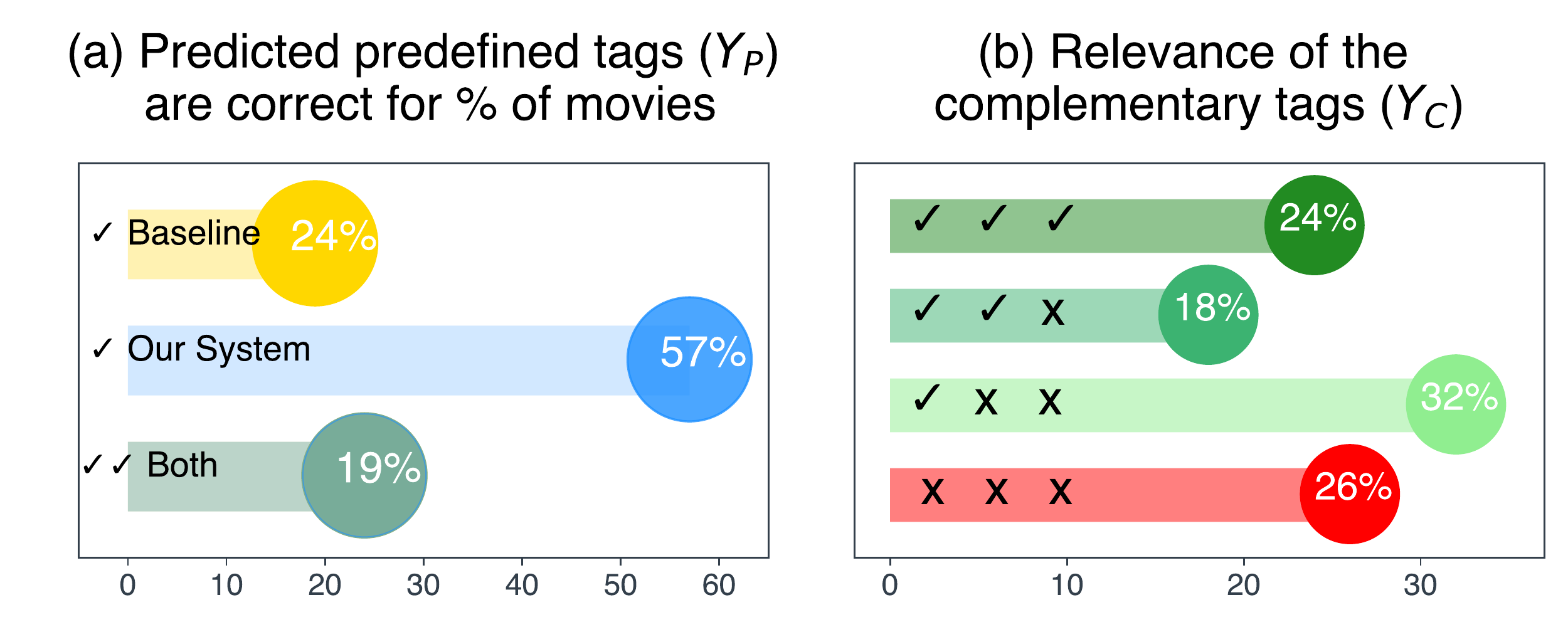}\\
    \includegraphics[width=0.9\columnwidth]{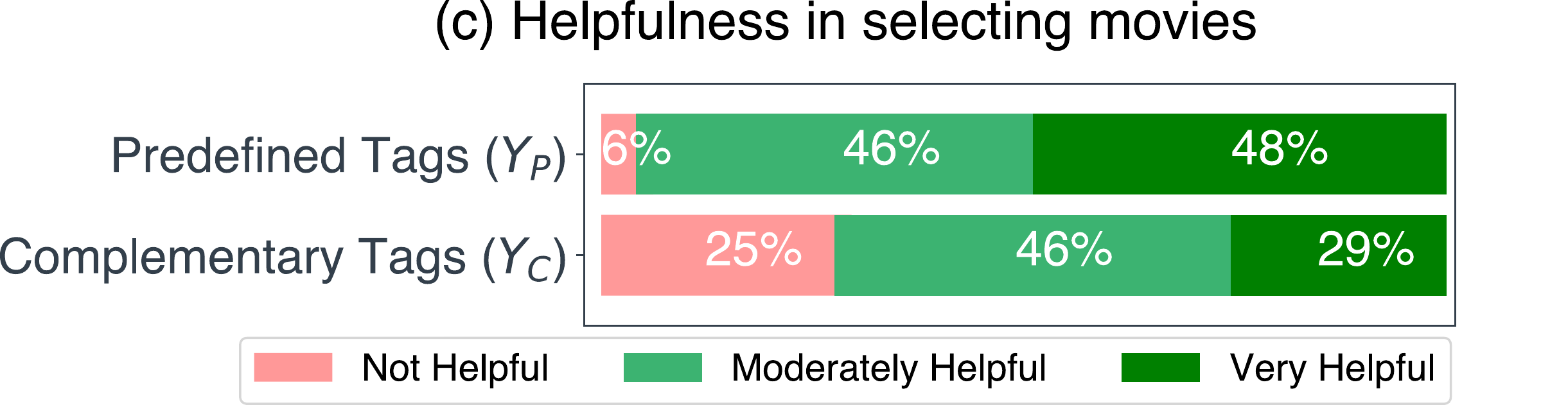}
    
    \caption{\small Summary of human evaluation results. (a) Comparing the correctness of two systems' predictions, (b) $\checkmark$ and $\times$ indicate rating from three human judges. e.g.,$\checkmark \checkmark \checkmark$: all judges marked 24\% complementary tags as correct, $\checkmark \checkmark \times$: two judges marked 18\% tags as valid, and so on. (c) Judges' feedback about whether our tagset helps users pick a movie by providing a quick description.}
    
    \label{fig:human_eval_pie}
\end{figure}

We perform a human evaluation experiment to verify the second research question, Q2 further, and answer Q3.
Additionally, we also want to investigate: ``\textit{how useful are the predicted tags from the predefined tagset ($Y_P$) and reviews ($Y_C$) for end-users to get a quick idea about a movie?}"

To explore Q2, we select \mbox{CNN-EF} \cite{folksonomication2018} as the baseline system\footnote{We used the online demo system released by the authors to generate the tags.}  to compare the quality of our tags for 21 randomly sampled movies from the test set.
For each movie, we instruct three human judges to read the synopsis to understand the story.
Then we show them two sets of tags for each movie and ask them to choose the tags that correctly describe the story.
In the first set of tags, we show only tags from $Y_P$, but we combine tags predicted by our model and those by the baseline system\footnote{If the tagsets from two systems are [\textit{a, b, c, d, e}] and [\textit{b, d, e, f, g}], we present [\textit{a, b, c, d, e, f, g}] and ask raters to select the correct ones. i.e., if \textit{a} is selected, System-1 gets one vote. If \textit{b} is selected, both systems get one vote.}.
In the second set of tags,  $Y_C$, we present the complementary tags extracted from the reviews (Section \ref{sec:review_tags}).
Figure \ref{fig:human_eval_pie}(a) shows that, for the predefined tags, our tagsets were more relevant than the baseline ones for 57\% movies, the baseline tags were better than \textit{HN(A)+MIL} for 24\% movies, and both systems were equally performing for 19\% movies.
Therefore, we get further verification of Q2. i.e., using reviews improves the retrieval of relevant tags from the predefined gold tagset.

To answer Q3, 141 open-vocabulary tags ($Y_C$) were rated by three judges.\footnote{114 distinct tags and $\approx$7 tags per movie.}
Figure \ref{fig:human_eval_pie}(b) shows that 24\% of these tags were rated relevant by all three judges, 18\% tags by two judges, and 32\% tags by one judge.
That means, $\approx$74\% of these tags were marked as relevant by at least one judge.

Finally, in Figure \ref{fig:human_eval_pie}(c), we assess the value of extracting tags to provide users a snapshot of the movie and make a \textit{go} or \textit{no go} decision on them. Results show that in 94\% of the cases, predicted tags from $Y_P$ were considered relevant in deciding whether to watch the movie or not. At the same time, in 75\% of the cases, complementary tags were also deemed relevant.

\subsection{Information from Reviews and Synopses}
\begin{figure}
    \centering
    \includegraphics[width=\columnwidth]{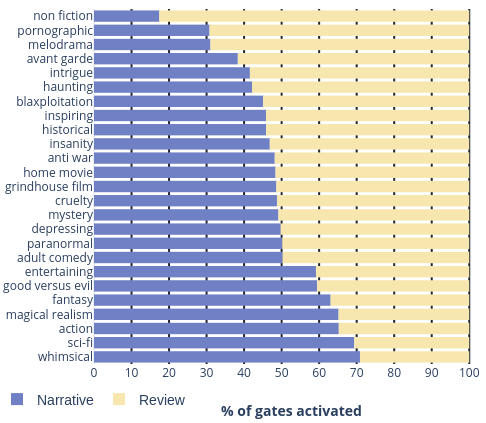}
    \caption{\small Percentage of gates activated ($z > 0.5$) for synopses and reviews. More active gates indicate more importance of the source for certain tags.}
    \label{fig:gate_activations}
\end{figure}

By analysing the predictions using only synopses and having user reviews as an additional view, we try to understand the contribution of each view in identifying story attributes.
We notice that using user reviews improved performance for tags like \textit{non fiction, inspiring, haunting}, and \textit{pornographic}. 
In Figure \ref{fig:gate_activations}, we observe that the percentage of activated gates for the reviews was higher compared to synopses for the instances having the mentioned tags.
Again, such tags are more likely to be related to visual experience or feeling that might be somewhat challenging for the model to understand only from written synopses.
For example, synopses are more important to characterize \textit{adult comedy} stories, but \textit{pornographic} representation can be better identified by the viewers and this information can be easily conveyed through their opinion in reviews.

\subsection{Generalization Capability}
In this section, we perform a few qualitative tests to assess the generalization capability of our model.
First, we observe the quality of our generated tags for some recently released movies that are not present in our dataset.
Finally, we check the quality of tags generated by our model for non-movie narratives, such as children's stories, ghost stories, novels, and TV series as it provides a scope to check if the model can generalize across domains.

\paragraph{Predictions for New Movies}
Back in 2019, we crawled plot synopses for a few recently released movies that did not have any tags at the time of collection. The goal of this experiment was to assess the quality of the tags predicted by our system for movies not in the train/dev/test set.

Table \ref{tab:new_movie_predictions} shows the predictions,  where we underline the tags that match user tags assigned since then.
For example, our predictions for {\fontfamily{ppl}{\selectfont \textit{The Irishman}}} are \textit{murder, neo noir, revenge, violence, flashback}, where most of these tags except \textit{neo noir} were found in IMDB.
Note that, accumulating reviews and tags is a time-consuming process, and many movies do not receive any reviews or tags at all.
We will check again in the coming months to see what tags appear for these movies. But this small-scale experiment bodes well with our previous results and our overall goal of automatically generating relevant tags from synopses.
\begin{table}[t]
    \centering
    \small
    \resizebox{\columnwidth}{!}{%
    \begin{tabular}{p{\columnwidth}}
    
    \cellcolor[HTML]{d8e8fe} \textbf{The Irishman: }  \underline{murder}, neo noir, \underline{revenge}, \underline{violence}, \underline{flashback}\\\addlinespace[0.15cm]
    
    \cellcolor{tagyellow!20}\textbf{Avengers Endgame: } \underline{good versus evil}, fantasy, \underline{action}, \underline{violence}, \underline{flashback}\\\addlinespace[0.15cm]
    
    \cellcolor[HTML]{d8e8fe}\textbf{Long Shot:} entertaining, \underline{comedy}, satire, humor, \underline{romantic}\\\addlinespace[0.15cm]
    
    \cellcolor{tagyellow!20}\textbf{Annabelle Comes Home: }  paranormal, \underline{horror}, gothic, cult, good versus evil\\\addlinespace[0.15cm]
    
    \cellcolor[HTML]{d8e8fe}\textbf{Once Upon a Time in Hollywood:}  comedy, \underline{violence}, \underline{cult}, \underline{humor}, \underline{murder}\\
    
    
    
    \end{tabular}}
    \caption{\small System predicted tags for movies released in 2019. The underlined tags match recently assigned tags from users in IMDb.}
    \label{tab:new_movie_predictions}
\end{table}

\paragraph{Generalization across Domains}
We also investigate the generalizability of our trained model. Instead of movie synopsis, we give as input a few popular stories from other domains like \textit{children stories}, \textit{modern ghost stories}, \textit{novels}, and \textit{TV series}\footnote{We collected the stories and synopses from the web. Sources with more examples are in Appendix \ref{app:out_of_domain_main}.}. 
Results in Table \ref{tab:out_of_domain_tags_mini} show that, our system can indeed predict tags that are very relevant to the new types of stories. Therefore, we conclude that our approach also shows great promise for other domains and can be extended with little effort.
\begin{table}
    \centering
    \small
    
    \resizebox{\columnwidth}{!}{
    \begin{tabular}{p{\columnwidth}}
        \multicolumn{1}{c}{\fontfamily{cmss}\textbf{Children Stories}} \\ 
         
            \cellcolor[HTML]{d8e8fe}\textbf{Cinderella:} fantasy, cute, romantic, whimsical, psychedelic\\ 

            \cellcolor{tagyellow!20}\textbf{Snow White and the Seven Dwarfs:} fantasy, psychedelic, romantic, good versus evil, whimsical\\  
        
        \addlinespace[0.15cm]
        \multicolumn{1}{c}{\fontfamily{cmss}\textbf{Modern Ghost Stories}}\\ 

            \cellcolor[HTML]{d8e8fe}\textbf{A Ghost:} haunting, flashback, atmospheric, murder, paranormal\\ 

            \cellcolor{tagyellow!20}\textbf{What Was It:} paranormal, haunting, gothic, horror, atmospheric\\ 
        
        \addlinespace[0.15cm]
        \multicolumn{1}{c}{\fontfamily{cmss}\textbf{Novels}}\\ 
        \cellcolor[HTML]{d8e8fe}\textbf{Romeo and Juliet:} {\fontfamily{cmss}revenge, murder, romantic, flashback, tragedy}\\ 

        \cellcolor{tagyellow!20}\textbf{The Hound of the Baskervilles:} murder, mystery, gothic, paranormal, flashback\\ 
 
        \addlinespace[0.15cm]
        \multicolumn{1}{c}{\fontfamily{cmss}\textbf{TV Series}}\\ 
            \cellcolor[HTML]{d8e8fe}\textbf{Game of Thrones S6E9: } violence, revenge, murder, action, cult\\ 
            \cellcolor{tagyellow!20}\textbf{Narcos Season 1:} murder, neo noir, violence, action, suspenseful\\ 

    \end{tabular}}
    \caption{\small Tags generated by our system for narratives that are not movie synopsis.}
    \label{tab:out_of_domain_tags_mini}
    \vspace{-1em}
\end{table}{}
    
    \section{Conclusion}
    In this paper, we focused on characterizing stories by generating tags from synopses and reviews.
We modeled the problem from the perspective of Multiple Instance Learning and developed a multi-view architecture.
Our model learns to predict tags by identifying salient sentences and words from synopses and reviews.
We demonstrated that exploiting user reviews can further improve performance and experimented with several methods for combining user reviews and synopses.
Finally, we developed an unsupervised technique to extract tags  that identify complementary attributes of movies from user reviews.
We believe that this coarse story understanding approach can be extended to longer stories, i.e., entire books, and are currently exploring this path in our ongoing work.

    \section*{Acknowledgments}
    
    We would like to thank the anonymous reviewers for providing helpful comments to improve the paper. We also thank Jason Ho for his helpful feedback on previous drafts of this work. This work was also partially supported by NSF grant 1462141. Lapata acknowledges the support of ERC (award number 681760, ``Translating Multiple Modalities into Text'').

    \bibliography{emnlp2020}
    \bibliographystyle{acl_natbib}
    
    \clearpage
    
    \appendix
    \clearpage
\section*{\centering Appendix}
\section{Data Pre-processing and Input Representation}\label{app_1}
We tokenize the synopses and reviews using \texttt{spaCy}\footnote{\url{http://spacy.io}} NLP library.
To remove rare words and other noise, we retain the words that appear at least in ten synopses and reviews ($<.01\%$ of the dataset).
Additionally, we replace the numbers with a \texttt{cc} token.
Through these steps we create a vocabulary of $\approx$42K word tokens.
We represent the out of vocabulary words with a \texttt{$<$UNK$>$} token.
For each movie sample, there are two text inputs (\textit{written synopsis} and \textit{summary of reviews}).
We use an empty string as the review text for the movies not having any review.

\section{Implementation and Training} 
We develop our experimental framework using PyTorch\footnote{\url{http://pytorch.org}}.
We use KL divergence as the loss function for the network and train the models for 50 epochs using Stochastic Gradient Descent (SGD) ($\eta=0.2$, $\rho=0.9$) as the optimizer.
We empirically set a dropout rate of 0.5 between the layers and $\ell_2$ regularization ($\lambda = 0.15$) to prevent overfitting.
We observe faster convergence using batch normalization after each layer.

\paragraph{Tuning Hyper-parameters}
During the experiments for developing the model, we
evaluate different model components using several combination of different hyper-parameters. Table \ref{tab:hyperparameter} presents the hyper-parameter space we explore.
While optimizing the model with Adam, we set the maximum number of epochs to 20, where the best validation performance were typically found after the 5th epoch with a learning rate of $1e^{-3}$.
Optimizing with SGD takes more epochs (typically around 30th epoch) even with high learning rates like $0.2$, but we observe better performance ($\approx 2\%$ higher MLR).

\begin{table}[!hbtp]
    \centering
    \begin{tabular}{|ll|}
    \hline
    \textbf{Hyper-parameter} & \textbf{Exploration Space}\\
    \hline
    RNN & LSTM$^*$, GRU  \\ \hline
    LSTM Units & 16, 32$^*$, 64, 128, 256\\\hline
    Optimizer & Adam, SGD$^*$\\\hline
    \multirow{3}{*}{$\eta$} & Adam : $1e^{[-4,-3^*, -2]}$ \\& SGD: $\mathrm{0.01, 0.05, 0.1,}$\\& $\mathrm{0.2^*, 0.3, 0.5}$\\\hline
    $\lambda$ & $\mathrm{0.005, 0.01, 0.15^*, 0.2}$\\\hline
    Dropout & $\mathrm{0.1, 0.2, 0.3, 0.4, 0.5^*}$\\ \hline
    
    Window context & $\mathrm{10, 20^*, 30, 40, 50}$\\
    \hline
    \end{tabular}
    \caption{\small Hyper-parameters and their values explored for tuning the model to achieve optimal performance on the validation data. $^*$ indicates the value providing the best performance.}
    \label{tab:hyperparameter}
\end{table}{}

\section{Results on Validation Set}\label{app:val_results}
Table \ref{tab:val_results} shows our results obtained on the validation set. We can see that our designed model outperforms all the baselines for predicting tags from only synopses. When we add review summaries, gated fusion performs best among the three aggregation methods, and it outperforms the system that only uses synopses.
\begin{table}
\centering
\resizebox{\columnwidth}{!}{%
\begin{tabular}{>{\collectcell\mymathrm}l<{\endcollectcell}
H
>{\collectcell\mymathrm}c<{\endcollectcell} >{\collectcell\mymathrm}c<{\endcollectcell}
>{\collectcell\mymathrm}c<{\endcollectcell} >{\collectcell\mymathrm}c<{\endcollectcell}}
\hline

& & \multicolumn{2}{c}{{$\mathrm{\mathbf{Top-3}}$}} & \multicolumn{2}{c}{\textbf{$\mathrm{\mathbf{Top-5}}$}}\\
\cmidrule(lr){3-4}
\cmidrule(lr){5-6}

\multicolumn{1}{c}{} & \multicolumn{1}{H}{{MLR}} &\multicolumn{1}{c}{$\mathrm{\mathbf{F1}}$} & \multicolumn{1}{c}{$\mathrm{\mathbf{TL}}$} & \multicolumn{1}{c}{$\mathrm{\mathbf{F1}}$} & \multicolumn{1}{c}{$\mathrm{\mathbf{TL}}$}\\ \hline
\multicolumn{6}{l}{Synopsis to Tags}\\ \hdashline
Most \quad Frequent  & 85.23 & 29.70 & 3 & 31.50 & 5\\
CNN-EF  & 86.10 & 37.70 &37 & 37.60 & 46\\ 
SBERT & 90.30 & 37.69 & 36 & 37.79 & 42\\

HN (Maxpool) & 89.36 & 36.72 & 17 & 36.19 & 28\\
HN (A) & 90.59 & 38.39 & 34 & 38.29 & 45\\
HN (A) + MIL & 91.32 & \mathbf{38.54} & \mathbf{49} & \mathbf{38.99} & \mathbf{54}\\

\hline 
\multicolumn{6}{l} {Synopsis + Review to Tags}\\ \hdashline
Merge\quad Texts & 92.63 & 41.52 & 54 & 41.17 & 61 \\
Concat\quad Representation & 92.81 & 41.61 & 56 & 41.46 & 64 \\
Gated\quad Fusion^* & 93.05 & \mathbf{41.65} & \mathbf{63} & \mathbf{42.05} & \mathbf{67} \\

\hdashline 
\multicolumn{6}{l}{Subset: Every movie has at least one review} \\ \hdashline
Synopsis & 91.01 & 38.49 & 47 & 38.68 & 54 \\
Review & 93.52 & 42.96 & 60 & \mathbf{43.29} & 65\\
Both & 93.52 & \mathbf{43.33} & \mathbf{64} & 43.02 & \mathbf{66}\\\hline

\end{tabular}}
\caption{\small Results obtained on the validation set using different methodologies on the synopses and after adding reviews with the synopses. TL stands for \textit{tags learned}. $^*$: t-test with $p$-value $< 0.01$.}
\label{tab:val_results}
\end{table}

\section{Out of Domain Stories}\label{app:out_of_domain_main}
In Table \ref{tab:chap_7_out_of_domain_tags}, we provide the tags generated for stories that are not movie synopsis. We also provide the source URL from where we collected the narratives. As our model is not suitable for handling very long texts like in novels, we collect their synopses and generate tags from those.
\begin{table*}[t]
    \centering
    
    \resizebox{0.9\textwidth}{!}{
    \begin{tabular}{p{0.5\textwidth}p{0.45\textwidth}}
        \toprule
         \textbf{Children Stories}& \\ \midrule
         
         Cinderella & {fantasy, cute, romantic, whimsical, psychedelic}\\
         \multicolumn{2}{l}{\scriptsize\url{https://www.thefablecottage.com/english/cinderella}}
         \\  \hline

        Snow White and the Seven Dwarfs & {fantasy, psychedelic, romantic, good versus evil, whimsical}\\ 
         \multicolumn{2}{p{0.6\textwidth}}{\scriptsize\url{https://www.storiestogrowby.org/story/snow-white-and-the-seven-dwarfs-bedtime-stories-for-kids/}}\\ \hline

        The Story of Rapunzel, A Brothers Grimm Fairy Tale &
        {fantasy, good versus evil, psychedelic, cute, gothic}\\ 
         \multicolumn{2}{p{0.6\textwidth}}{\scriptsize \url{https://www.storiestogrowby.org/story/early-reader-rapunzel-fairy-tale-story-kids/}}\\\hline
        
        The Frog Prince: The Story of the Princess and the Frog & {fantasy, cute, whimsical, entertaining, romantic}\\
         \multicolumn{2}{p{0.6\textwidth}}{\scriptsize\url{https://www.storiestogrowby.org/story/princess-and-the-frog-story-bedtime-stories-for-kids/}}\\
         \hline
        
        Aladdin and the Magic Lamp from The Arabian Nights & {fantasy, good versus evil, action, romantic, whimsical}\\ 
         \multicolumn{2}{p{0.6\textwidth}}{\scriptsize\url{https://www.storiestogrowby.org/story/aladdin-story-from-the-arabian-nights-bedtime-stories-folk-tales-for-kids/}}\\
        \hline
        
 \midrule
 
 \textbf{Modern Ghost Stories}\\ 
 \multicolumn{2}{p{0.7\textwidth}}{\scriptsize\url{https://www.gutenberg.org/files/15143/15143-h/15143-h.htm}}\\
 \midrule
 The Shadows on The Wall by Mary E. Wilkins Freeman &
{haunting, gothic,  murder, horror, atmospheric}\\\hline

The Mass of Shadows By Anatole France &
 {fantasy, atmospheric, gothic, murder, romantic}\\ \hline

A Ghost By Guy De Maupassant &
{haunting, flashback, atmospheric, murder, paranormal}\\ \hline

What Was It? By Fitz-James O'Brien &
{paranormal, haunting, gothic, horror, atmospheric}\\ \hline

 \midrule
 \textbf{Novel Summaries}\\
 \midrule
 Romeo and Juliet by William Shakespeare & {revenge, murder, romantic, flashback, tragedy}\\
\multicolumn{2}{p{0.8\textwidth}}{\scriptsize\url{ https://www.booksummary.net/romeo-and-juliet-william-shakespeare}}\\ \hline

Harry Potter and Sorcerer's Stone by J. K. Rowling & {fantasy, good versus evil , entertaining, action, comedy}\\ 
\multicolumn{2}{p{0.8\textwidth}}{\scriptsize\url{https://www.booksummary.net/harry-potter-and-the-sorcerers-stone-j-k-rowling}}\\ \hline

Oliver Twist by Charles Dickens & {murder, revenge, flashback, romantic, violence}\\ 
\multicolumn{2}{p{0.8\textwidth}}{\scriptsize\url{https://www.booksummary.net/oliver-twist-charles-dickens}}\\\hline

The Hound of the Baskervilles by Arthur Conan Doyle & {murder, mystery, gothic, paranormal, flashback}\\ 
\multicolumn{2}{p{0.8\textwidth}}{\scriptsize\url{https://www.booksummary.net/hound-of-baskervilles-arthur-conan-doyle}}\\
\hline
 
 \textbf{TV Series}\\
 \midrule
 Game of Thrones Season 6 Episode 9: Battle of the Bastards & {violence, revenge, murder, action, cult}\\
\multicolumn{2}{p{0.8\textwidth}}{\scriptsize\url{https://en.wikipedia.org/wiki/Battle_of_the_Bastards}}\\ \hline
        
    The Big Bang Theory Season 3 Episode 22:
The Staircase Implementation &
    {romantic, comedy, flashback, entertaining, psychedelic}\\
    \multicolumn{2}{p{0.8\textwidth}}{\scriptsize\url{https://www.imdb.com/title/tt1648756/plotsummary}}\\\hline
    
    Friends Season 5 Episode 14: The One Where Everybody Finds Out &
 {comedy, entertaining, adult comedy, humor, romantic}\\
 \multicolumn{2}{p{0.8\textwidth}}{\scriptsize\url{https://en.wikipedia.org/wiki/The_One_Where_Everybody_Finds_Out}}\\\hline

    Narcos Season 1 &
  {murder, neo noir, violence, action, suspenseful}\\ 
  \multicolumn{2}{p{0.8\textwidth}}{\scriptsize\url{https://en.wikipedia.org/wiki/Narcos_(season_1)}}\\\hline

    \bottomrule
    \end{tabular}}
    \caption{Example of stories that are not movie synopsis and the source URL. Tags in the right column are generated by our system.}
    \label{tab:chap_7_out_of_domain_tags}
\end{table*}{}


\onecolumn
\section{Examples of Plot Synopses and Reviews}\label{app:example_plot_review}

\begin{longtable}[c]{p{0.95 \textwidth}}
\toprule
\multicolumn{1}{c}{\textbf{Rush Hour 3}}\\
\hline

\multicolumn{1}{c}{\textbf{Plot Synopsis}} \\
\fontfamily{cmss}\scriptsize Three years after the end of Rush Hour 2, James Carter is no longer a detective, but a traffic cop on the streets of Los Angeles. Lee is now the bodyguard for his friend Ambassador Han, the former Consul from the first film. Lee is still upset with Carter about an incident in New York City when Carter accidentally but not fatally shot Lee`s ex-girlfriend in the neck, Secret Service agent Isabella Molina.During the World Criminal Court discussions, as the Ambassador addresses the importance to fight the Triad, he announces that he may know the whereabouts of Shy Shen. 
Suddenly, Han takes a bullet in the shoulder, disrupting the conference. Lee pursues the assassin and corners him, discovering that the assassin is his "brother" Kenji. When Lee hesitates to shoot Kenji, Carter shows up driving towards the two and accidentally nearly runs Lee over, allowing Kenji to escape.In the hospital, Lee learns that Han will make a full recovery. Han's daughter, Soo Yung (Zhang Jingchu), now grown up, arrives and makes Lee and Carter promise to capture the one behind the shooting. She then informs Lee and Carter that her father gave her an envelope which contains important information regarding the Triad, and that the envelope is in her locker at the martial arts studio where she works. Lee and Carter make their way to the martial arts studio where they battle a giant, (Sun Ming Ming), but find out that a gang of armed men had already arrived and taken the contents from the locker. Lee and Carter are told by the Master of the studio that Soo Yung and Han are in danger and rush back to the hospital.Once the two reach the hospital, a gang of assassins arrive to kill Soo Yung and Han. Lee and Carter manage to defeat them, with the help of Soo Yung, and interrogate the leader of the assassin squad. Much to Lee and Carter's surprise, the Asian assassin only speaks French. With the help of a resident nun, Sister Agnes (Dana Ivey), in translation, they find out that they are marked for death by the Triad along with Soo Yung and Han. For her protection, they take her to the French Embassy and leave her under the care of Reynard, the French ambassador. When a car bomb detonates, nearly killing Reynard and Soo Yung, Lee and Carter decide to go to Paris to investigate.In Paris, (after getting a painful cavity search from a Parisian commissioner, played by Polanski) Lee and Carter meet up with George, a taxi driver. George refuses to drive Carter, saying that Americans make him sick, as they are "the most violent people on Earth"to which Carter replies by forcing George, at gunpoint, to drive to a Triad hideout disguised as a gentleman's club. There Lee fights off a Triad assassin named Jasmine (Youki Kudoh); meanwhile, Carter meets a beautiful woman whose name is not disclosed(Noémie Lenoir). However, Lee and Carter are both forced out of the club and are captured by the Triads. Lee and Carter manage to escape, but then have a falling out concerning Lee's relationship with Kenji. Shortly after Carter leaves, Reynard appears. Lee asks who Shy Shen is, but Reynard tells him that Shy Shen is not a man, but a list of the Triad leaders. Reynard reveals that Han's informant knows where Shy Shen is. The informant turns out to be Geneviève, the woman Carter met the gentleman's club and both Lee and Carter end up looking for her.After the two have encountered Geneviève they save her from an assassination attempt by the Triads and flee to their hotel room.[3] However, they are attacked again by Jasmine and decide to hide out with George, who now has fostered a great appreciation for the United States. Lee and Carter learn that Geneviève not only knows where the list is, she is the list. The names of the thirteen Triad leaders have been tattooed on the back of her head, as per tradition, and Genevieve explains that she will be decapitated and buried if the Triads capture her. When Lee and Carter bring Geneviève to Reynard, he asks Geneviève to show him the list. Lee points out that they never told him that she was the list. Reynard then reveals that he has been working with the Triads all along. Kenji calls and informs Lee that he has captured Soo Yung and that he would like to exchange Soo Yung for Geneviève.Lee arrives at the exchange point, the Jules Verne Restaurant in the Eiffel Tower, with Carter, disguised as Geneviève. During a sword fight, Lee and Kenji fall off the tower and get caught in a safety net.
Kenji's sword cuts the safety net open and it collapses, leaving both men hanging on for dear life. Lee then grabs Kenji's arm, intending to save his life. Kenji then willingly lets go of Lee and presumably falls to his death, saving Lee's life, who then, with less weight, swung to a scaffolding. Meanwhile, Carter single-handedly defeats the rest of the Triad henchmen, unwittingly kills Jasmine, and saves Soo Yung. As they send Soo Yung down the elevator, more Triads arrive. In order to escape, Lee and Carter use a French flag as a makeshift parachute and float to safety. Unfortunately, they are confronted by Reynard, who is holding Geneviève hostage and threatening to kill her and frame Lee and Carter for her murder. However, George, having followed Lee and Carter, manages to shoot Reynard and declares "Case Closed". The police suddenly arrive, with the commissioner from earlier gloating and trying to get undeserved credit. After giving the commissioner a team punch to the face, Lee and Carter leave the scene dancing to War.\\ 

\multicolumn{1}{c}{\textbf{Review Summary}}\\
\fontfamily{cmss}\scriptsize Not only that, but this movie seemed more rushed and the story wasn't as well developed as the first two Rush Hour films.In the beginning it seems like Lee and Carter are not on good terms, Carter seems to have broken off Lee's relationship to his ex girlfriend, Isabella.
But I think if you loved the first two Rush Hour films, you should see this, it's still a fun film and has great comedy, there is a little less action, a warning in advance, but it's all good if you are looking for a fun film for the end of the summer.6/10.
The first two were pretty good comedy movies staring both Jackie Chan and Chris Tucker as cop partners with a typical plot with good action mixed with comedy.
There's quite a few decent action scenes and again Chris Tucker's character delivers a lot of the jokes and Jackie Chan's still got the awesome moves and stunts but is it just me or was the action a bit tamer compared to the other two?
So anyway, you've got your B level action-comedy movie and it does it quite well.Jackie Chan (Who plays as Lee) acts well as usual (have some problem for pronouncing things but I didn't mind that) and Chris Tucker is still hilarious and he's still loud as ever.
The action scene were always a thrill and the car chase was very fun (there was a bit of laughter in the theater).There's a lot more comedy (which some jokes were cheesy) in this movie than action, but in the end I didn't mind, I still felt entertained and happy to spend my 15\$ in this movie.So this is the end of my simple review, go watch this movie with low expectation and in a mood for some good laughter and action and you'll enjoy it.
Director Brett Ratner returns to his element, offering a third fun, funny, and violent slapstick installment in the Rush Hour series.Rush Hour 3 sees Inspector Lee and Detective Carter back together again, trying to save the lives of a Chinese ambassador who may be on the brink of cracking open a massive organized crime syndicate called the Triads, the ambassador's daughter, and a woman with a very dangerous secret.
Chris Tucker gives his best performance in the series, delivering a lot of comic punch and playing a nice complement to Chan's sombre and serious Inspector Lee.Rush Hour 3 delivers exactly what fans of the series look for - a simple linear action-thriller liberally decorated with a lot of cleverly written comedy and the amazing physical performance of Jackie Chan.
The original Rush Hour was a Jackie Chan vehicle of sorts to break into Hollywood, and it made a lot of money with the mis-pairing opposite Chris Tucker in a buddy cop movie formula filled with action and comedy.
The action sequences in Rush Hour 3 look a bit tired, tame, and very uninspiring, and what Jackie Chan can probably still do, has been whittled down to sequences that are just a pale shadow of what could have been.Which leaves us with the comedy, thankfully still having its moments especially for those punchlines which deliver.
If there's any hint of rudimentary character development after these years, is that his James Carter, besides having been relegated to traffic duties, managed to "half-chinese" himself, and no longer is that helpless cop who without his gun, can't kick a ball for nuts.The plot is no rocket science, and in fact, the previous two movies just had something which could coast along from scene to scene, providing a platform either to get our heroes Lee (Chan) and his brother-from-another-mother James (Tucker) into fisticuffs, or provide something for laughs.
Some of the more totally insane moments involve those deliberate lost in translation moments, which are the more enjoyable moments in this movie.Rush Hour 3 is similar in structure with its predecessors, and it does seem a tad familiar at times in the way the story gets developed, with only a change in locale, now set in Paris.
No prizes for guessing the other two.It's not that Rush Hour 3 is a particularly bad movie, but it's just a tired re-working of themes already heavily mined in the first two instalments. The only\\ \fontfamily{cmss}\scriptsize marginal difference here is that whereas in Episode 1 and 2 the "player-outside-his-turf" mantle was worn alternately by Lee (Jackie Chan) and Carter (Chris Tucker) respectively, here, they're both out of their comfort zone viz.
Europe, and more specifically, Paris This plot point doesn't actually change the formula's outcome that much, because every other element is essentially unchanged.The coolest stuff in the movie is already showcased in the trailer, so if you're pressed for time, just watch that.PS: Roman Polanski did look like he was having fun though...so much fun, in fact, that he didn't bother to have his name added to the credits.
And Chan, the straight-laced, 'good guy' Inspector Lee. The charisma is still there, just like it was since the first Rush Hour movie.
OK, here are a few reasons why I disliked (note: NOT hated) this film.Chan looks bored and his action scenes simply do not impress.Tucker gets a few decent one-liners but surprisingly, his comic timing is off.Storyline is simultaneously boring and unbelievable, and the ending a massive letdown.Everyone seems to be in simply for the money, and there's clearly ZERO imagination or honesty visible.
They both seem to be "just in it for the money"I really hope this will be the last Rush Hour film, the first and second was both really really enjoyable with fun comments, good comedy and good action scenes this third installment has nothing of it.For some good Jackie action stick with his older films and those made in Hong Kong, for Chris tucker films watch any other film he's in besides this one..The goofy antics of Chris Tucker mixed with the amazing martial arts moves of Jackie Chan made both of the previous Rush Hour films funny and entertaining.
I admit it that it is funny and some jokes really make me laugh but now, we are focusing on the story plot which is straight-forward and its action scenes are not very impressive for a Jackie Chan movie.
Okay, I've not seen either of the first two Rush Hour films, so don't have that direct yardstick to compare them to, but I have seen a good number of action comedies over the years and this one added absolutely nothing; in fact, it was damned close to reducing my enjoyment of the whole genre.
It has so much going on but is understandable, no hard story lines to follow, it flows nicely and will make you think what a good film, this is sure to bring more rush hour's i hope, can't wait till the next installment!
Here we have another example of a film franchise that is too successful for its own good, and in the year of threequels, this was one we didn't need.Set a few years after Rush Hour 2, James Carter (Chris Tucker) is no more then an LA traffic cop and Lee (Jackie Chan) is the head bodyguard for the Chinese Ambassador.
James and Lee end up working together again, and take their investigation Paris and set out to break-up the Triad.The plot is skeleton thin, like they have been in the previous Rush Hour films; but this is a new low!
A couple funny lines in a movie doesn't make it a funny movie.Second if you watch the movie close enough it copy's, takes or borrows plot devices and scenes from a lot of other movies that were way better.I want people to think about this \$140 MILLION dollars(just let that sit there for a second) Third Since when did a Jackie Chan movie become so filled with special effects that the stunts don't even seem the least bit real therefore you can't even become engaged in the action.
Very mixed, but these were the general thoughts..."Not enough action.", "A Chris Tucker movie.", "Same jokes used.", etc.I want to think that most of Rush Hour 3's problems are attributed to the fact that it was just a rushed film.
The first two films are not that much longer, Rush Hour 2 is also 90 minutes, but they had good paces and weren't lazy productions.As for what some of the other reviews say, I say this: Jackie Chan is getting old, which may be the reason why the film didn't seem to have as much action.
I think that Tucker may have gotten more screen time, however...and I will say he was still pretty damn funny.However, the true laughs are brought on by Yvan Attal, who, as their French cab driver, is inspired by the violence taking place around him to become an, "American who kills for no reason." I am an American and I thought that was hilarious.The bottom line is this: Rush Hour 3 is lazy, rushed, and not as good as the first two.
Its a Mark of a quality film that makes you feel sad thats it ended and so quickly especially when it is 90 minutes in length.the chemistry between these two great stars in still there and the gags come thick and fast and have everybody in stitches.they both are putting some age and weight on but that does not stop the stars from delivering punch after punch whether thats as a joke or to the adversaries.my favourite gag was the one they have used throughout the series with chris tuckers confusion over the words and names me and you or who.
Matters aren't helped by the fact that some jokes can be very hit or miss and the storyline is so obvious it unreal, I swear if you can't figure out who the villain is in the first few minutes then you are a blithering idiot.Part of the charm of the first two Rush Hour movies was the chemistry between Jackie Chan and Chris Tucker.
Rush Hour 3 follows the adventure of detectives Carter and Lee (played by Chris Tucker and Jackie Chan) as they follow an important case into Paris.
They battle impossible foes doing equally impossible feats of daring duo all while being sprite, charming and funny.In the end, they come out smelling like roses.The end.Rumors of filming a fourth Rush Hour film immediately after this third installment were apparently scrapped due to unknown reasons, but one could easily venture that failing interest in the series hamstrung such a project (but who knows ...maybe there will be a fourth).Tucker and Chan still have a unique chemistry on-screen, one a speed talking black man and the other a soft talking Chinese gentleman.
Sure, there were some fun points here and there, Roman Polanski delivered a cute cameo and the Parisian settings were a nice change of scenery for the more adrenaline rushed aspects of the story - but as a whole, I felt this just wasn't enough for a franchise this big.I wanted to like it more, particularly after being pleasantly surprised by many of the other sequels this year (Pirates 3, Spider-Man 3, Evan Almighty, Live Free and Die Hard and Ocean's Thirteen to be precise), but it just didn't work for me in the end.I gave this a disappointed 5/10..
The problem with the 3rd of the series is that in every way it feels very forced, it also feels like they agreed to make the film so they could all work together again.Nothing in the movie feels fresh, or exciting, and nothing in the movie helps enhance the first two instalments.The Good.....Not much actually, Jackie Chan can still do some quite cool, quite difficult choreographed fight scenes, even though we have all seem much better now (Casino Royale anyone?) I actually can't think of anything else positive I will take away from this movie....The Bad.....Without contradiction, the fight scenes.
Directed by Brett Ratner, who helmed the first two installments in the series, the film sends the pair to Paris in pursuit of the assassin, where they become involved with crime lords, the French police, a helpful taxi driver, Jackie's long lost brother, and some beautiful women.As expected in a Jackie Chan movie, the action is a non-stop series of martial arts fights, car chases, and explosions.
While there is nothing ground-breaking here, the film has plenty of action for Jackie's fans, and some good lines from Tucker; when you throw in appearances by Max Von Sydow, Philip Baker Hall, and Roman Polanski, "Rush Hour 3" is pretty decent entertainment for the third film in a series..
Nonetheless, Rush Hour 3 provides just enough laughs and fun action to be an enjoyable watch.It follows the story of the attempted assassination of Ambassador Chan and the following chase of the Triads that did it, along with the protection of a woman with knowledge of the organisation's secrets by the Blackenese duo Lee and Carter.
Chris Tucker is just as charismatically black as in the previous Rush Hour films and Jackie Chan is exactly the same as usual – we would expect no less and no more — and yes, charismatically is a new word, I'm basically Shakespeare.
It still manages to be a pretty good action comedy flick with just enough Jackie Chan and Chris Tucker jokage to keep it at a relatively high quality among other films of its kind..
However, there are sequels that are not as good as the previous offering, and then there are God-awful, art-less, money-making piles of turd...very much like Rush Hour 3.As an audience, we've more or less had enough of Chris Tucker by the end of Rush Hour 2, but take a look at his CV, and you'll see that the poor guy couldn't get any work after the second movie, so along came Jackie Chan to help his mate out, allowing him to reprise his role as quite possibly the most annoying man on the planet.Brett Ratner too has effectively directed his own demise, with a distinct lack of vision, and what can only be described as sheer contempt for his audience.The film is set in France, yet French people speak English to one another, like it's an everyday occurrence, using Jewish insults like, "schmuck," to enhance the Gallic feel of the overall piece.
This is the third film that director Rather worked with Tucker and Chan on the previous two "Rush Hour" movies.
Imagine this movie without Chris Tucker and the jokes and you will have a dull Chinese movie with no artistic value.The acting is horrific, the story line is childish and almost cartoon like; the plot is so predictable they didn't bother thinking about making a change from the original 2 movies.Jacky Chan is clearly older and not capable to doing the same amazing stunts that brought him to fame in Rush Hour 1 \& 2 (10 years ago), other than that I don't think he can act..the only saving grace for me is the Hilarious Carter who at times reminded me of American Dad the cartoon with his off the cuff impulsive comments..
I loved the first two films so i thought another sequel would just be 90 minutes more of action comedy enjoyment, i went to see the first showing in my local cinema and was stunned at how little fight scenes and unimpressive fights there were, which brought me to only a few conclusions why this was.1)American film makers don't let their stars do as much stuff as they used to.2)Brett Ratner decided Jackie C couldn't help choreograph fight scenes.Or god forbid age has finally caught up with Jackie which until he brings out another movie seems to be the logical explanation, now i will say Jackie Chan has done excellently in the past also bearing in mind Jackie is 55 years old so it has to happen eventually.I did find the script entertaining there were some funny moments and loud over the top remarks coming from Cris Tucker which i was expecting, i think as a family film it would be good but for classic Jackie Chan fans they'l be disappointed.over all i give Rush Hour 3, 5 out 0f 10 7 out of 10 for a funny script 2.5 out of 10 for action p.s I felt a similar disappointment with Die Hard 4.0 i just hope Stallone dosn't do the same with Rambo 4..
The subplot between Inspector Lee and his Brother was awkward and i thought it was very fake and un-believable.But we don't go to see a Rush Hour film for it's story, but rather for the comedy that Chris Tucker brings and the fighting that Jackie Chan brings.
His scenes were great and easily the funniest scenes in the movie, in my opinion.I can't say i can recommend this film to anyone but die hard Rush Hour fans or fans of Chris/Jackie.
The only people that are going to see it are going to be the fans of the first two Rush Hours, which, though they did get progressively worse, were still enjoyable action-comedy flicks with a healthy balance of Jackie Chan cracking ribs and Chris Tucker wise-cracking.
Another sequel, of another movie that i was a fan off, and this this time with an actor who I've liked since when i can remember.Jackie Chan may not have delivered path breaking cinema or enjoy the iconic or cult status that Bruce Lee got, but what he has always delivered has been good fun, with the right blend of humour, action, and naughtiness.
Six years after Rush Hour 2 comes this sequel which further chronicles the misadventures of mismatched cops Inspector Carter (Chris Tucker) and Inspector Lee (Jackie Chan).\\
\\
\midrule
\multicolumn{1}{c}{\textbf{August Rush}}\\ \hline
\multicolumn{1}{c}{\textbf{Plot Synopsis}} \\
\fontfamily{cmss}\scriptsize  In 1995, Lyla Novacek (Keri Russell) is a cellist studying at the Juilliard School and living under strict rule of her father (William Sadler). Louis Connelly (Jonathan Rhys Meyers) is the lead singer of "The Connelly Brothers", an Irish rock band. Lyla and Louis meet at a party after their respective concerts, and sleep together on the rooftop under a full moon, to the music of a street performer below. The day after, they separate in a hurry, and are unable to maintain contact as Lyla is ushered away by her father to Chicago. Later, Lyla realizes she is pregnant, and after an argument with her father, she is struck by a car. Due to the accident trauma, she gives birth prematurely, and her father secretly puts the baby boy up for adoption, allowing her to believe that her son died.
Eleven years later, Evan Taylor (Freddie Highmore) is living in a boys' orphanage outside New York City, where he meets Richard Jeffries (Terrence Howard), a social worker with Child and Family Services. Evan has the savant-like ability to hear music wherever he is, making him a bullying target for the older orphans. Convinced that his parents will find him, Evan runs away to New York City, "following the music" in the hope it will lead him to his family. He finds a boy named Arthur (Leon Thomas III) performing in Washington Square Park.
Louis, who left the band the same night Lyla was struck, now lives in San Francisco as an agent, while Lyla has also given up performing and now lives in Chicago teaching music. Louis reconnects with his brothers at a birthday party for one of the other band members, and after an argument and fistfight over breaking up the band, he decides to reconnect with the woman he now knows is Lyla. Lyla is called to her father's deathbed, where he confesses that her son is alive and in New York, since her father believed that he was only doing it for both him and his daughter and that that her son could've destroy her future. Lyla abandons her father to his fate and heads to New York to look for him.
Evan follows Arthur to his home in a condemned theater, and is taken in by Maxwell "Wizard" Wallace (Robin Williams), a vagrant, arrogant, abusive and aggressive musician who teaches homeless children music and employs them as street performers. Evan tries playing Wizard's prize guitar, Roxanne, and is so good that Wizard gives him the guitar and his old spot in Washington Square Park (both of which were previously Arthur's). He gives Evan the stage name "August Rush" and tries to market him to clubs. Seeing the posters that Jeffries has posted for the runaway Evan, Wizard destroys all the ones he finds, hoping to keep Evan and his gift for his own gain.
On arriving at Lyla's apartment in Chicago, Louis talks to one of her neighbors, who mistakenly tells Louis she's on her honeymoon. Despairing, he ends up in New York, where he gets his band back together. After Jeffries meets Wizard and Arthur on the street and becomes suspicious, the police raid the derelict theatre in which Wizard and his "children" are living.
Wizard helps Evan evade the police, telling him never to reveal his real name to anyone. Evan (now "August") takes refuge in a church where a young girl, Hope (Jamia Simone Nash), introduces him to the piano and written music. He picks up both skills so quickly that Hope gets the attention of the parish pastor (Mykelti Williamson), who takes August to Juilliard where he once again impresses the faculty. A rhapsody takes shape from August's notes and homework.
In New York, Lyla goes to Jeffries' office, and Jeffries identifies Evan as her son. While looking for him, she takes up the cello again and accepts an offer to perform with the Philharmonic at a series of concerts in Central Park. August is selected to perform the rhapsody he's been composing at the same concert. However, Wizard, who found out about August's performance by Arthur, interrupts the rehearsal and claims to be his father, and manages to pull August out of the school. On the day of the outdoor concert, August is back in his spot in Washington Square, while Wizard makes plans to smuggle him around the country to play. He meets Louis and, unaware of their blood relationship, they have an impromptu guitar duet. August tells him of his dilemma, and Louis encourages him to go. That evening, with help from Arthur, August escapes from Wizard through the subway and heads for his concert. Louis, after his own performance with his reunited band, sees Lyla's name on one of the concert banners and also heads for the park. Jeffries finds a misplaced flyer for "August Rush" with a picture, and realizing August is Evan, also heads for the concert.
August arrives in time to conduct his rhapsody, which attracts both Lyla and Louis to the audience, where they are reunited. August finishes his rhapsody and as he turns to discover his parents, he smiles knowing that he has been right all along.\\

\multicolumn{1}{c}{\textbf{Review Summary}}\\
\fontfamily{cmss}\scriptsize I was abandoned a second time by one of my parents.The movie "August Rush" was healing to my soul wounded since early childhood;and again, in my early twenties.
Instead of music, I used my talent of writing to deal with the lost of my parents."August Rush" made me fantasize during the movie that my yearning and searching for my parents were like this remarkable child..
The scenes have an appealing fantasy element, while at the same time, the plot manages to explore true-to-life human situations such as bullying of those who are different.The music is incredible, and mostly consists of original scores.
It includes gospel, rock and classical, seamlessly integrated in a new way that works extremely well.The plot is somewhat predictable and possibly a little "sappy", but those elements are easily overcome by the moment-to-moment execution of the story.
Think of a modernized "Oliver" with Robin Williams as Fagin to a group of homeless, musically talented kids...plus extra elements of romance and intrigue, and you will have a bit of an idea about this movie.The three main characters are all physically "beautiful" people who manage to convey the story with a minimum of dialog.
In the end, this movie is at least an endorsement and celebration of the significance of music in our lives and at most a transcendent, fun experience to watch.I rarely like to see any movie more than once, but definitely want to see this again.
There were some errors but for the most part I applaud the film makers for the attention to the musical details.Yes the movie was a bit bit corny, and a little over the top, but for the most part I loved it and suggested it to every one..
Although at times, there may seem like there are gaps in the story line/character development, the point of this is that the music is what communicates those hidden details of the movie.Overall, this film is a masterpiece that should be cherished by music-lovers everywhere..
Ridiculously laughable story, hammy, bad acting, sub par music, zero chemistry between the two romantic leads, sticky sweet, implausible plot are only a few of the ways I can describe this incredibly bad movie.
I don't want to give anything away, in case for some insane reason you want to see it, so I'll only go into the story problems in the first 15 minutes.The actors were trying their hearts out--except Robin Williams, whom I normally love but found false and unsatisfying--and in some cases were able to overcome the material and give fairly good performances(Terence Howard, Kerri Russell).
While the viewer knows what will eventually happen, there is still much film to go...during which you see Evan's amazing talents burst forth AND an evil manipulator, 'the Wizard' (Robin Williams, sees the boy's potential and takes him in and puts him to work.
Robin Williams seems to be making a different movie than everyone else.Look, we're not cynics; we love "Love, Actually," "About a Boy," and all of Frank Capra, but the story has to seduce you in, not knock you to your knees; has to have a level of believability that doesn't require you to swallow logs when straining at gnats.This was a dud.
With a star cast that consisted of Robin Williams, Terence Howard, Jonathan Rhys Meyers and one of the biggest star kids in Hollywood -Freddie Highmore, a story about a musical genius, it has to be a brilliant movie, hasn't it?
And since the movie revolves around music and the fact that August is a "prodigy", more effort should have been put into teaching the kid to look like he knew anything about music or conducting.
People who have never considered nor been exposed to the processes behind music might not notice a problem, but to those who have, the film's central character will more closely resemble a comic-book-superhero version of a musician than any musician, genius or no, who has ever lived.This young lad's extraordinary ability (to reach professionalism at any instrument, and even theory/notation, within seconds of coming into contact with it for the first time) is only a symptom of the problem.
I don't remember the last time I saw a Hollywood film which genuinely felt like it was the vision of an artist who really had something to say.Perhaps this film actually was written by someone who loves and understands music and wanted to convey something about it, but the fact that you can't tell simply by watching the film is a testament to its failure..
The father says the kid died but really he had given the boy to child services.Meanwhile, the boy - now 11 and living in an orphanage - runs off to New York City somehow figuring that magically, through hearing music, he'll meet his parents.
It weakly conveys that the main character (a boy named Evan Taylor) believes he has some kind of control over the field, but it falls short of the intended correlation to his sensation of "music everywhere" which is better illustrated later in the film.
Instead, my friends, I am going to tell you to please, use some of your time watching this movie, so you can come back here and rate it with this one little painful star, as I did.
Com'on, ye people of sense and sensibility, join me and suffer this movie and be merry you found the strength and then give it the only rating it deserves....PS: I am not going to elaborate on the "oh-you're-so-mean!" side characters and the sanctified trinity of the main characters and their ridiculously hollow "deep" emotions within their hearts of jelly who have no backbone to ever efficiently stand up for themselves and all they do is cry and moan and dream under the moon and hope that their sorry lives will be solved by cosmical deliverance because they sure won't do a thing since they are made of childish fairy-jelly glow, ...nor on the plot and dialogue silliness, and such, ...because you can read all about it on other reviews.\\
\fontfamily{cmss}\scriptsize  The story is taken directly from Oliver Twist, but the actual events, it isn't even believable in the fairytale context that it is told.The central love story is based on a night of rooftop sex with strangers who have about 3 minutes of the corniest lines I could possibly come up with before going at it like guinea pigs in heat: Who are you Lyla?" She pauses, smiles then looks off in the distance,
"I'm just.. me." And it just goes on and on like that.The rest of the movie is filled with similar vomit inducing dialogue; I don't know how many times "You've got to believe in your music" was repeated.Most of the characters are cardboard cutouts: There's rich cellist girl with overbearing father, brooding punk rocker, inner city black kid who talks just like you'd expect, and perpetually dazed and confused skinny white kid who gets picked on but always follows his dream.The cinematography is mostly boring and standard, what you might see in a car commercial or something similarly mundane.
A lighter touch in directing, and better musical direction would have turned this into a good film - the story is interesting, but it was painted with a jackhammer instead of a brush..
I rented August Rush a few days ago with good expectations, I just heard wonderful things about this film, so I was excited to see it.
We also have the talented Jonathan Rhys Meyers from the Tudors, the incredible Terrence Howard, then a disturbing yet memorable performance from Robin Williams.Evan Taylor is an orphan who is just convinced that he can hear his parent's music, that they do want him and he goes out to find them.
Evan is set to find his parents but come across a musical group of kids, who are making money for Wizzard, when Evan plays music, it's magical, Wizzard exposes Evan giving him a new name, August Rush and makes money off of him.
But when Lyla finds out about her son being alive, she goes to New York to find him, Louis starts thinking about Lyla and finds out where she lives, Chicago, but when he finds out she's gone, he goes to New York to relive his band days, instead they find each other and the musical genius their love created.August Rush is one of the first films of the year that I'm rating a perfect 10, because there is nothing wrong with this film, to be honest, I think it deserves a higher rating than a 7.4.
August Rush is a movie that I'm sure will work it's way into your heart, it's a magical film that is absolutely perfect.10/10.
He believes that he will find his parents if they only hear his music, because they are truly bonded by their music.(My Comment) The movie is a human interest story about a young boy's unyielding faith and will power to never give up on his dream.
Mozart would be an absolute imbecile compared to this little kid August Rush, and for those familiar with music, this aspect (the foundation, really) just kills the movie.It is impossible to play like Michael Hedges in your first few minutes with a guitar.
However, I just finished watching August Rush and I am in no way exaggerating when I say that it is by far the best movie I have ever seen.
It not only grabs your heart from the very first scene but it grabs your soul within the first thirty minutes and by the time the movie's climax arrives you're in it so deep that whether you're Mr. Macho who wouldn't cry at his own mother's funeral or just someone who's bored and wants to see something that will be worth watching you will undoubtedly be wiping your eyes with your shirt sleeve and not caring who sees because everyone around you will be wetting their sleeves as well.
I know its a musical, but i think listening to ARR songs is a better resort to watching this incredibly stupid sappy cheesy movie.3/10.
This is one of the most ridiculous movies I've watched in recent years.Essentially its the story of an orphan who is trying to find his parents and does so through music.It comes across like a fairytale, but even Alice In Wonderland was more realistic than this.
Im not sure who would actually enjoy this movie, maybe if you're 70, or under 12 but for everyone else I'd save your time.The acting itself wasn't bad, though the more interesting characters were played by Terrence Howard and Robin Williams, and they were both severely under-developed as you wanted to know more about them and less about this kid with the stupid smile all the time..
However, while the movie has a modern setting, it shares many plot elements with OLIVER TWIST, ending even better.It begins with a young couple of musicians that meets and has a one-night stand, and when she becomes pregnant her dad does everything to make her believe that the child died at birth, although he just put the child for adoption.A decade later the boy, Evan, lives in a orphanage and is mocked by the other kids because of his talents in music, that makes him like a savant with powerful skills.
In the meanwhile an inexplicable series of circumstances draw his mother and dad that search for their son (after they discovered he is NOT dead), but also Evan ends in New York, becoming a star for his talents in music while a evil man, Michael Wallace known as ''the Wizard'' (Robin Williams in a rare villanious role) takes advantage of the boy and wants to make money with his talents.
The story touched my heart that I nearly cried...I think that "August Rush" should get Oscar's for at least music.
Such a music I can listen for hours and days and it ever makes me bored off it.Seriously I can say that the first time I saw preview, I knew that it will be maybe the best movie during my life and actually it is.
If you are a true music fan, musician, or even love soundtracks in the sense of Instrumentals, then this film will be a hit with you.No film, in years of my life have touched me the way this film has, how has this film gone under the radar for so long?
The movie tells a story of a orphan boy who is a musical genius and who believes that his parents will find one day .
Imagine a modern version of Oliver Twist, add beautiful classical and rock music and finish the mix with a somewhat predictable, but nevertheless intriguing love story and "August Rush" is the best that you can possible make out of it.By watching the opening scene, showing Freddie Highmore in a corn field, I already suspected this could become something very special.
I have never seen or heard anything like it and I sincerely hope that I'll one day watch a movie that does an even better job.
Perhaps it's better to categorize this movie as being targeted at people of all ages, that love music and want to see a feel-good movie with beautiful music and top notch acting.
I was hoping for him and moved by the music he felt in the world around him.Normally I don't watch movies of this genre, but August Rush captured my emotions.
Howard is a fine actor as shown earlier this year in The Brave One (see my review) but he's not given one of the bigger and more important roles and not a lot to work with so he kind of gets out shined.I am not a musical expert nor a musician of any kind, much to my disappointment so they could have played those instruments in the film with their feet and it couldn't have mattered to me.
You will either buy into this contrived tale of orphan who hears music then runs off New York to find his parents or you won't.I didn't.Cloyingly cute I had no patience for it and squirmed from the opening bits of the kid conducting the wind, through the flash back meeting of young rock star dad and rich girl cellist torn apart by her disapproving dad, on to the really annoying Robin "Please put my head in a duck press to make me stop screaming" Williams and then to the oh gee ending.
Although this film plays well to a broad audience, it is very mystical and based on simple, yet emotional themes that will play flat to some movie-goers.If you have strong parental feelings or enjoy movies centered on the power of human love and attraction, this story will move you like few films ever have.
However, if you are easily bored with themes that are lacking in danger and suspense or prefer gritty true-to-life movies, this one may come off as a disappointment.The screenplay seems written as a spiritual message intimating that there is an energy field that connects all of life, and music is one of the domains available to any who care to experience it.The plot is simple but deep in implication- an orphaned boy wants to reunite with his parents and feels that his inherited musical genius can somehow guarantee their return.
I was not impressed.The acting was unconvincing, the character development nonexistent: Russell, who spends most of the film gazing wistfully at various points on the horizon, evokes virtually no sympathy as a woman who loses her son and lapses into a decade-long depression; Rhys-Meyers is flat and uninspiring as August's tortured rock-musician-turned-bond-trader father ("I'm sad...I'm Irish...I'm sad"); and Williams is only mildly creepy as Wizard, the Fagin-like ringmaster of a coterie of musical ragamuffins who live in (of course!) an abandoned theater.
August Rush is quite simply one of the best movies I have seen in years.
This was one of the best movies I have seen in a very long time, maybe ever...The music, the dialogue, the story, it all speaks right to the heart and it leaves you glowing with happiness and admiration.
I thought the premise of the story was a good one; A boy given up for adoption attempts to find his birth parents, because his love for music gives him the sense that his parents are still alive.
I'm not a fan of musical movies; however, I thought I would give this film a try.
Once I watched the kid Freddie Highmore tearing for the loss of his parents I was starting to get attached to the movie, and once I started tasting the music this film contains and with all beautiful emotions flying around I was amazed and could not take my eyes off the screen till the finish..
This movie has some great acting it makes you believe like it's an actual story.
Overall this movie is great for both the family and also for anyone interested in Music, it was truly inspirational, I loved it, it was everything I expected and more, its definitely underrated..
If the story intrigues you, see it, I'm pretty sure you'll feel good too:-).August Rush is worth watching for us who still have some dreams inside of us and don't need to be reminded of the world we live in all the time!.
'Evan Taylor' AKA August Rush (Freddie Highmore) is placed in an orphanage, longing for parents he believes he can 'hear' in the music of the spheres.
There's only a few movies a year that you watch and you just don't want it to end..
Amazing movie with an incredible story about a little boy who never gave up and kept faith.I watched this several times with my mom, she loves it, and it is an excellent family movie.
August Rush is a must watch film because it's a very inspirational story.Remarkable movie filled with love, art, music, inspiration, and hope.
\\ \bottomrule

     \caption{Examples of plot synopsis and review summary for some movies.}
    \label{tab:my_label}
\end{longtable}
\newpage
\begingroup
\section{Detail Results of the Human Evaluation}\label{app:human_eval_table}
\begin{longtable}[c]{p{0.19 \textwidth}  p{0.73 \textwidth}}
    \toprule
    
    \textbf{IMDb id and Title} & \textbf{Tags with Number of Votes}\\ \midrule
    
\makecell[l]{ tt0401398  \\ \small Kronk's New Groove \\  
\small{ $[$\textcolor{blue}{\faMehO} \textcolor{teal}{\faSmileO} \textcolor{teal}{\faSmileO}$]$
$[$\textcolor{red}{\faFrownO} \textcolor{blue}{\faMehO} \textcolor{blue}{\faMehO}$]$}}
& \makecell[l]{
\textbf{B}: feel-good$^3$, magical realism$^3$, cute$^2$, whimsical$^1$, prank$^1$
 \\ 
\textbf{N}: flashback$^3$, romantic$^3$, entertaining$^2$, prank$^1$, psychedelic
 \\ 
\textbf{R}: kronk$^2$, moralising$^2$, funny$^2$, cartoon$^2$, hilarious$^1$, gags$^1$
} \\ \midrule

\makecell[l]{tt0112887 \\ \small The Doom Generation\\ $\spadesuit$ \small{ $[$\textcolor{blue}{\faMehO} \textcolor{blue}{\faMehO} \textcolor{blue}{\faMehO}$]$
$[$\textcolor{blue}{\faMehO} \textcolor{teal}{\faSmileO} \textcolor{teal}{\faSmileO}$]$}} & \makecell[l]{
\textbf{B}: pornographic$^3$, adult comedy$^1$, neo noir$^1$, comic, blaxploitation
 \\ 
\textbf{N}: violence$^3$, murder$^3$, pornographic$^3$, sadist$^2$, cult$^1$
 \\ 
\textbf{R}: disturbing$^3$, deranged$^3$, tortured$^3$, erotic$^3$, violent$^3$, psychotic$^3$, goriest$^3$, \\ sexual$^3$, kinky$^2$, weirdness$^2$, nihilistic$^1$, gay$^1$, homoerotic$^1$, irony, \\humourous, goth
} \\ \midrule

\makecell[l]{tt0780606 \\ \small Shock to the System \\  \small{ $[$\textcolor{red}{\faFrownO} \textcolor{teal}{\faSmileO} \textcolor{teal}{\faSmileO}$]$
$[$\textcolor{teal}{\faSmileO} \textcolor{blue}{\faMehO} \textcolor{blue}{\faMehO}$]$}}  & \makecell[l]{
\textbf{B}: plot twist$^3$, murder$^3$, suspenseful$^2$, intrigue$^2$, neo noir
 \\ 
\textbf{N}: murder$^3$, queer$^3$, plot twist$^3$, flashback$^2$, romantic$^1$
 \\ 
\textbf{R}: whodunit$^3$, lesbian$^3$, lgbt$^3$, gay$^3$, vengeance$^3$
} \\ \midrule

\makecell[l]{tt0239949\\ \small Say It Isn't So \\ $\clubsuit$ \small{ $[$\textcolor{teal}{\faSmileO} \textcolor{teal}{\faSmileO} \textcolor{teal}{\faSmileO}$]$
$[$\textcolor{blue}{\faMehO} \textcolor{teal}{\faSmileO} \textcolor{red}{\faFrownO}$]$}
}  & \makecell[l]{
\textbf{B}: comedy$^3$, adult comedy$^3$, humor$^2$, dramatic$^2$, entertaining$^1$
 \\ 
\textbf{N}: comedy$^3$, romantic$^2$, humor$^2$, prank$^1$, entertaining$^1$
 \\ 
\textbf{R}: humour$^3$, funny$^1$
} \\ \midrule

\makecell[l]{tt0083869\\ \small Eating Raoul \\ $\spadesuit$  \small{ $[$\textcolor{blue}{\faMehO} \textcolor{teal}{\faSmileO} \textcolor{teal}{\faSmileO}$]$
$[$\textcolor{blue}{\faMehO} \textcolor{blue}{\faMehO} \textcolor{teal}{\faSmileO}$]$}
} & \makecell[l]{
\textbf{B}: neo noir$^3$, adult comedy$^2$, humor$^1$, comedy$^1$, bleak$^1$
 \\ 
\textbf{N}: murder$^3$, adult comedy$^2$, pornographic$^2$, satire$^1$, comedy$^1$
 \\ 
\textbf{R}: violent$^3$, slapstick$^2$, humour$^2$, masochistic$^1$, bondage, kinky
} \\ \midrule

\makecell[l]{tt0109650\\ \small Doomsday Gun \\ $\spadesuit$  \small{ $[$\textcolor{teal}{\faSmileO} \textcolor{blue}{\faMehO} \textcolor{teal}{\faSmileO}$]$
$[$\textcolor{blue}{\faMehO} \textcolor{red}{\faFrownO} \textcolor{teal}{\faSmileO}$]$}} & \makecell[l]{
\textbf{B}: dramatic$^3$, historical$^2$, suspenseful$^1$, thought-provoking$^1$, neo noir$^1$
 \\ 
\textbf{N}: violence$^3$, intrigue$^3$, murder$^2$, flashback, alternate history
 \\ 
\textbf{R}: thriller$^3$, cynical$^2$, backstabbing$^1$, conspiracy$^1$, amusing$^1$, evil$^1$, chases$^1$, \\ paradox$^1$, nightmare$^1$, doomsday$^1$, chilling, mi6
} \\ \midrule

\makecell[l]{tt0373283\\ \small Saints and Soldiers \\ \small{ $[$\textcolor{teal}{\faSmileO} \textcolor{blue}{\faMehO} \textcolor{teal}{\faSmileO}$]$
$[$\textcolor{red}{\faFrownO} \textcolor{red}{\faFrownO} \textcolor{blue}{\faMehO}$]$}}  & \makecell[l]{
\textbf{B}: historical$^3$, action$^3$, dramatic$^3$, suspenseful$^1$, realism
 \\ 
\textbf{N}: violence$^3$, historical$^3$, murder$^2$, suspenseful$^1$, flashback
 \\ 
\textbf{R}: massacre$^3$, brutality$^2$, affirming$^1$, brotherhood, underbelly, christianity
} \\ \midrule

\makecell[l]{tt0191423\\ \small Scooby-Doo \\ \small Goes Hollywood \\ $\clubsuit$ \small{ $[$\textcolor{red}{\faFrownO} \textcolor{red}{\faFrownO} \textcolor{blue}{\faMehO}$]$
$[$\textcolor{blue}{\faMehO} \textcolor{blue}{\faMehO} \textcolor{teal}{\faSmileO}$]$}}  & \makecell[l]{
\textbf{B}: entertaining$^2$, humor$^1$, comic$^1$, psychedelic, horror
 \\ 
\textbf{N}: cult$^1$, flashback$^1$, comic$^1$, psychedelic, horror
 \\ 
\textbf{R}: scooby$^3$
} \\ \midrule

\makecell[l]{tt0175059\\ \small Power Rangers Lost\\ \small Galaxy\\ $\clubsuit$ \small{ $[$\textcolor{teal}{\faSmileO} \textcolor{blue}{\faMehO} \textcolor{blue}{\faMehO}$]$
$[$\textcolor{red}{\faFrownO} \textcolor{blue}{\faMehO} \textcolor{blue}{\faMehO}$]$}}  & \makecell[l]{
\textbf{B}: good versus evil$^3$, sci-fi$^2$, fantasy$^2$, alternate history$^1$, comic
 \\ 
\textbf{N}: good versus evil$^3$, fantasy$^2$, violence$^1$, paranormal$^1$, psychedelic
 \\ 
\textbf{R}: mystical$^2$, mythic$^1$, cartoon, psycho, magical, funny
} \\ \midrule

\makecell[l]{tt0088805\\ \small The Big Snit \\ $\spadesuit$  \small{ $[$\textcolor{blue}{\faMehO} \textcolor{blue}{\faMehO} \textcolor{teal}{\faSmileO}$]$
$[$\textcolor{teal}{\faSmileO} \textcolor{teal}{\faSmileO} \textcolor{blue}{\faMehO}$]$}}  & \makecell[l]{
\textbf{B}: thought-provoking$^1$, suspenseful$^1$, comic$^1$, paranormal, bleak
 \\ 
\textbf{N}: psychedelic$^3$, absurd$^2$, cult$^1$, philosophical$^1$, satire
 \\ 
\textbf{R}: surreal$^3$, absurdist$^2$, existential$^2$, cartoon$^1$, demented
} \\ \midrule

\makecell[l]{tt0064072\\ \small Battle of Britain \\  \small{ $[$\textcolor{teal}{\faSmileO} \textcolor{teal}{\faSmileO} \textcolor{teal}{\faSmileO}$]$
$[$\textcolor{teal}{\faSmileO} \textcolor{teal}{\faSmileO} \textcolor{red}{\faFrownO}$]$}}  & \makecell[l]{
\textbf{B}: historical$^3$, action$^3$, dramatic$^2$, thought-provoking$^1$, anti war
 \\ 
\textbf{N}: historical$^3$, flashback$^2$, violence$^2$, anti war, suspenseful
 \\ 
\textbf{R}: gripping$^3$, tragic$^2$, biographical$^2$, dogfights$^1$, sixties$^1$
} \\ \midrule

\makecell[l]{tt0067500\\ \small La noche del terror ciego \\ $\clubsuit$  \small{ $[$\textcolor{teal}{\faSmileO} \textcolor{teal}{\faSmileO} \textcolor{teal}{\faSmileO}$]$
$[$\textcolor{blue}{\faMehO} \textcolor{red}{\faFrownO} \textcolor{blue}{\faMehO}$]$}}  & \makecell[l]{
\textbf{B}: suspenseful$^2$, paranormal$^2$, murder$^2$, violence$^2$, revenge$^1$
 \\ 
\textbf{N}: violence$^2$, murder$^2$, cruelty$^2$, cult$^1$, flashback$^1$
 \\ 
\textbf{R}: disturbing$^3$, satanic$^1$, gore$^1$, eroticism, lesbianism, visions, torture, tinged,\\ subversive
} \\ \midrule

\makecell[l]{tt0117913\\ \small A Time to Kill \\ $\spadesuit$  \small{ $[$\textcolor{teal}{\faSmileO} \textcolor{teal}{\faSmileO} \textcolor{blue}{\faMehO}$]$
$[$\textcolor{teal}{\faSmileO} \textcolor{blue}{\faMehO} \textcolor{blue}{\faMehO}$]$}}  & \makecell[l]{
\textbf{B}: revenge$^3$, suspenseful$^2$, murder$^2$, violence$^2$, neo noir
 \\ 
\textbf{N}: revenge$^3$, murder$^2$, violence$^2$, flashback$^2$, sadist$^2$
 \\ 
\textbf{R}: violent$^3$, crime$^3$, brutally$^3$, vengeance$^2$, vigilante$^2$, sadism$^1$, poetic,\\ depraved, fictional
} \\ \midrule

\makecell[l]{tt0335345\\ \small The Passion of the Christ \\ $\spadesuit$ \small{ $[$\textcolor{teal}{\faSmileO} \textcolor{blue}{\faMehO} \textcolor{blue}{\faMehO}$]$
$[$\textcolor{teal}{\faSmileO} \textcolor{red}{\faFrownO} \textcolor{teal}{\faSmileO}$]$}}  & \makecell[l]{
\textbf{B}: dramatic$^3$, thought-provoking$^2$, historical$^2$, suspenseful$^1$, allegory$^1$
 \\ 
\textbf{N}: violence$^3$, christian film$^3$, murder$^2$, flashback$^2$, avant garde$^1$
 \\ 
\textbf{R}: brutality$^3$, symbolism$^3$, slasher$^2$, treachery$^2$, enlightening$^2$, torture$^2$, lucid$^1$,\\ occult$^1$, allusion$^1$, ironic
} \\ \midrule

\makecell[l]{tt1185616\\ \small Vals Im Bashir  \\ $\spadesuit$  \small{ $[$\textcolor{teal}{\faSmileO} \textcolor{blue}{\faMehO} \textcolor{teal}{\faSmileO}$]$
$[$\textcolor{teal}{\faSmileO} \textcolor{red}{\faFrownO} \textcolor{red}{\faFrownO}$]$}}  & \makecell[l]{
\textbf{B}: historical$^2$, thought-provoking$^2$, anti war$^1$, philosophical$^1$, alternate history
 \\ 
\textbf{N}: flashback$^3$, violence$^2$, storytelling$^2$, murder$^1$, psychedelic
 \\ 
\textbf{R}: nightmares$^3$, nightmare$^3$, surreal$^1$, escapist$^1$, surrealism$^1$, disturbing, witty
} \\ \midrule

\makecell[l]{tt2379386\\ \small \small In the Name of the King: \\ \small The Last Mission \\ $\clubsuit$ \small{ $[$\textcolor{blue}{\faMehO} \textcolor{teal}{\faSmileO} \textcolor{blue}{\faMehO}$]$
$[$\textcolor{red}{\faFrownO} \textcolor{blue}{\faMehO} \textcolor{blue}{\faMehO}$]$}}  & \makecell[l]{
\textbf{B}: action$^3$, fantasy$^3$, violence$^3$, good versus evil$^2$, historical fiction$^1$
 \\ 
\textbf{N}: violence$^3$, murder$^3$, good versus evil$^2$, revenge$^1$, flashback$^1$
 \\ 
\textbf{R}: antihero$^3$, magical$^3$, campiness$^1$, dungeon$^1$, rampage$^1$, cinematic$^1$,\\ masterpiece
} \\ \midrule

\makecell[l]{tt0085412\\ \small Deal of the Century \\ $\spadesuit$ \small{ $[$\textcolor{red}{\faFrownO} \textcolor{blue}{\faMehO} \textcolor{blue}{\faMehO}$]$
$[$\textcolor{red}{\faFrownO} \textcolor{teal}{\faSmileO} \textcolor{blue}{\faMehO}$]$}
}  & \makecell[l]{
\textbf{B}: dramatic$^3$, suicidal$^1$, realism$^1$, humor, thought-provoking
 \\ 
\textbf{N}: absurd$^3$, comedy$^2$, satire$^1$, cult$^1$, humor
 \\ 
\textbf{R}: maniacal$^3$, pathos$^1$, symbolism$^1$
} \\ \midrule

\makecell[l]{tt1355627\\ \small Evil Bong 2: King Bong \\ $\spadesuit$\small{ $[$\textcolor{blue}{\faMehO} \textcolor{blue}{\faMehO} \textcolor{blue}{\faMehO}$]$
$[$\textcolor{blue}{\faMehO} \textcolor{blue}{\faMehO} \textcolor{blue}{\faMehO}$]$}}  & \makecell[l]{
\textbf{B}: humor$^2$, clever$^1$, action$^1$, comic$^1$, thought-provoking
 \\ 
\textbf{N}: cult$^2$, comedy$^2$, violence$^1$, murder$^1$, revenge
 \\ 
\textbf{R}: humour$^2$, wicked$^2$, amusing$^1$, killer$^1$, evil$^1$, geeky$^1$, titular$^1$, laced,\\ irreverence, homophobic
} \\ \midrule

\makecell[l]{tt0023921\\ \small Cross Fire \\ $\spadesuit$ \small{ $[$\textcolor{teal}{\faSmileO} \textcolor{teal}{\faSmileO} \textcolor{teal}{\faSmileO}$]$
$[$\textcolor{red}{\faFrownO} \textcolor{blue}{\faMehO} \textcolor{red}{\faFrownO}$]$}}  & \makecell[l]{
\textbf{B}: suspenseful$^3$, murder$^3$, revenge$^2$, sadist$^1$, neo noir
 \\ 
\textbf{N}: murder$^3$, violence$^3$, suspenseful$^3$, revenge$^2$, flashback
 \\ 
\textbf{R}: gunfight$^3$, fistfights$^1$, classic
} \\ \midrule

\makecell[l]{tt0154749\\ \small Kudrat\\  $\spadesuit$ \small{
$[$\textcolor{blue}{\faMehO} \textcolor{blue}{\faMehO} \textcolor{blue}{\faMehO}$]$
$[$\textcolor{blue}{\faMehO} \textcolor{blue}{\faMehO} \textcolor{red}{\faFrownO}$]$
}}  & \makecell[l]{
\textbf{B}: melodrama$^2$, romantic$^2$, flashback$^2$, intrigue$^1$, paranormal$^1$
 \\ 
\textbf{N}: murder$^2$, flashback$^2$, romantic$^2$, revenge$^2$, paranormal$^1$
 \\ 
\textbf{R}: thriller$^3$, nightmares$^2$, reincarnation$^2$, chilling$^1$, karz, melancholy
} \\ \midrule

\makecell[l]{tt0098575 \\ \small Valmont\\ $\spadesuit$ \small{ $[$\textcolor{blue}{\faMehO} \textcolor{blue}{\faMehO} \textcolor{blue}{\faMehO}$]$
$[$\textcolor{red}{\faFrownO} \textcolor{blue}{\faMehO} \textcolor{red}{\faFrownO}$]$}}  & \makecell[l]{
\textbf{B}: romantic$^3$, melodrama$^2$, historical fiction$^1$, queer, intrigue
 \\ 
\textbf{N}: romantic$^3$, revenge$^2$, murder$^2$, violence$^2$, flashback
 \\ 
\textbf{R}: cynicism$^3$, irony$^2$, cruel$^1$, liaisons$^1$, humour$^1$, brutality, ruthless
} \\ \bottomrule

    \caption{\small Data from the human evaluation experiment. \textbf{B} represents the tags predicted by the baseline system, \textbf{N} represents the tags predicted by our new system, and \textbf{R} represents the open set tags extracted from the user reviews by our system. 
    If a tag is followed by a number in superscript, the number indicates the number of annotators who selected the tag as relevant to the story. 
    We consider a tag as relevant if it has at least two votes.
    $\spadesuit$ indicates the instances where our system's predictions were more relevant compared to the baseline system, and $\clubsuit$ indicates the opposite.
    For the rest of the instances, both systems had a tie.
    Annotators' feedback about the helpfulness of the tagsets (closed set tags and open set tags) are presented by emoticons ( \textcolor{teal}{\faSmileO}: Very helpful, \textcolor{blue}{\faMehO}: Moderately helpful, \textcolor{red}{\faFrownO}: Not helpful).  
    First three emoticons are the feedback from all the annotators for the tags from the baseline system and our system.
    Rest of the three emoticons are the feedback for the tags extracted from the user reviews.
    }
    \label{tab:huma_eval_results}
\end{longtable}
\endgroup
    
    \end{document}